%% file: main.tex
\documentclass[preprint,12pt]{elsarticle}

\usepackage{graphicx}   
\usepackage{subfig}     
\usepackage{booktabs}
\usepackage{amssymb}
\usepackage{amsmath}
\usepackage{url}
\usepackage{breakurl}
\usepackage[hidelinks]{hyperref}

\journal{Energy and Buildings}

\begin{document}

\begin{frontmatter}

\title{BuildSTG: A Multi-building Energy Load Forecasting Method using Spatio-Temporal Graph Neural Network}

\author[1,2]{Yongzheng Liu}
\author[3]{Yiming Wang}
\author[3]{Po Xu}
\author[4]{Yingjie Xue}
\author[2,5]{Yuntian Chen \corref{cor1}}
\ead{ychen@eitech.edu.cn}
\author[2,5]{Dongxiao Zhang \corref{cor1}}

\address[1]{The Hong Kong University of Science and Technology (Guangzhou), 511453, Guangzhou, China}
\address[2]{Ningbo Institute of Digital Twin, Eastern Institute of Technology, Ningbo, 315200, China}
\address[3]{Ginlong Technologies Co., Ltd., Ningbo, 315712, China}
\address[4]{Wuhan University, Wuhan, 430072, China}
\address[5]{Zhejiang Key Laboratory of Industrial Intelligence and Digital Twin, Eastern Institute of Technology, Ningbo, 315200, China}

\cortext[cor1]{Corresponding author}

\input{00_abstract}


\end{frontmatter}


\input{01_introduction}
\input{02_related_work}
\input{03_methodology}
\input{04_results}
\input{05_conclusions}
\input{06_Acknowledgment}

\bibliographystyle{elsarticle-num-names} 
\bibliography{ref}

\input{08_Appendix}

\end{document}

%% file: 00_abstract.tex
\begin{abstract}
Due to the extensive retention of building operation data, data-driven building load prediction methods have demonstrated powerful capabilities in forecasting building energy loads. Buildings with similar operating conditions, physical characteristics, and types often exhibit similar energy usage patterns, which are reflected in their operation data showing similar trends and spatial dependencies. However, conventional building load prediction methods have significant limitations in extracting these spatial dependencies. To address this challenge, this paper proposes a multi-building load prediction method based on spatio-temporal graph neural networks, which is divided into three main steps: graph representation, graph learning, and method interpretation. First, a graph representation method is developed that identifies building correlations based on intrinsic characteristics and environmental factors. This method constructs a graph by comparing energy consumption patterns across buildings. Next, a multi-level spatiotemporal graph convolutional architecture with an attention mechanism is designed to predict energy loads for multiple buildings. Finally, a model interpretation method based on the optimal graph structure obtained from the training process is developed. Experiments on the Building Data Genome Project 2 dataset validate that the proposed method outperforms commonly used baseline models like XGBoost, SVR, FCNN, GRU, and Naïve in terms of prediction accuracy. Additionally, the model demonstrates strong robustness and generalization, performing reliably under uncertainty and unseen data. Visualization of the building similarity matrix confirms the model’s interpretability, revealing its ability to group similar buildings and establish meaningful spatial dependencies, proving that the proposed Att-GCN method for learning spatial dependencies between buildings with similar energy usage patterns is both reasonable and interpretable.
\end{abstract}

%% file: 01_introduction.tex
\section{Introduction}
With urbanization increasing, building energy consumption and carbon emissions are growing. Construction and operation of buildings account for 34\% of global energy use, with 30\% from operations. They are responsible for 37\% of global carbon emissions, with 27\% from operations \cite{globalreports}. Therefore, energy conservation during the operational phase of buildings holds paramount importance for achieving energy efficiency and carbon reduction goals. Accurate prediction of building loads serves as a critical foundation for energy management tasks like architectural optimization design \cite{li2019ann}, system optimization control \cite{ilbeigi2020prediction}, demand side response, and energy audits. It is also the basis for optimization of building energy systems and fault diagnosis, facilitating efficiency enhancements.

Building energy consumption refers to the energy consumed from external sources during the operation of the building, including energy for maintaining the building’s environment and energy for activities within the building. Predicting building energy consumption typically involves forecasting cooling, heating, and electrical loads. Building energy consumption prediction methods can be divided into two types: physical-modeling based methods and data-driven methods. Physical-modeling based methods depend on heat transfer, thermodynamics, and HVAC (Heating, Ventilation, and Air Conditioning) domain knowledge to simulate buildings. They predict building energy consumption by simulating activities of people inside the building, equipment operation, and heat transfer processes between the interior and exterior of the building. These methods, based on physical domain knowledge, have interpretability and high reliability. However, they often necessitate detailed building information and expert knowledge, and the modeling process is time-consuming and labor-intensive, limiting their large-scale applicability. With the popularization of building automation systems, a large amount of operational data has been accumulated, fostering the advancement of data-driven building load prediction methods \cite{zhang2020hybrid}. Compared to physics-based methods, data-driven methods offer superior accuracy and modeling convenience, presenting significant potential for practical application \cite{gassar2020energy}. 

Building operational data exhibits strong temporal dependencies. Temporally, there is thermal inertia in buildings inherently, and the load data as a time series reflects periodic properties and has time-lag effect. Therefore, most data-driven methods for building load forecasting often focus on historical information only, utilizing previous load data to establish models. For instance, Tan et al. \cite{tan2024short} introduced a forecasting method combining SVMD algorithm and improved Informer model and applied it to the heat load forecasting of district heating system. Gao et al. \cite{gao2024hybrid} proposed a hybrid forecasting model (BAS-GRNN and LSTM) combing generalized regression neural network (BAS-GRNN) and long short-term memory neural network (BAS-LSTM) optimized by beetle antennae search algorithm for building cooling load prediction. Fan et al. \cite{fan2017short} compared the performance of various data-driven algorithms (MLR, RF, XGBoost, FCNN, SVR, etc.) in predicting building cooling load, finding XGBoost has the best performance. Wang et al. \cite{wang2021building} proposed a deep convolutional neural network based on ResNet for hour-ahead building load forecasting, which also significantly improved the prediction accuracy through feature fusion techniques. Bian et al. \cite{bian2022research} proposed a model incorporating time cumulative effects and an improved time convolution network (TCN) for power load prediction. Addressing multi-step prediction issues, Jung et al. \cite{jung2021attention} introduced a load prediction method utilizing an attention-based GRU network, which can reflect the previous point well to predict the current point. Bashir et al. \cite{bashir2022short} processed original data using Prophet and LSTM, and input the processed data into BPNN to enhance load prediction accuracy.  

However, it's crucial to note that building operational data also possesses distinct characteristics in spatial dimensions. In the spatial dimension, buildings with similarities in aspects such as type, physical attributes, geographical location, building age, and the number and type of occupants often demonstrate similar energy consumption patterns. This similarity is reflected in their operational data, as evidenced in prior studies that have shown spatial and typological correlations in energy use behaviors among buildings \cite{BERARDI2017230}\cite{GUPTA2023112920}. Nevertheless, the above-mentioned data-driven methods only focus on temporal features and treat each building as a single entity for prediction, often neglecting the spatial dependencies of buildings. In the spatial dimension, conventional approaches typically model buildings in isolation, relying solely on the historical operational data of a single building to train an energy load prediction model specifically that is only suitable for that building. These methods fail to fully leverage the spatial relationships between the target building and other buildings. This flaw leads to insufficient prediction accuracy and poor generalization ability, thereby reducing the reliability of energy-saving strategies based on load prediction, such as optimization control \cite{saeedi2019robust} and fault diagnosis \cite{zhang2020generic}.

In response to the constraints of traditional data-driven approaches in leveraging building spatio-temporal data for load prediction, researchers have proposed using graph neural networks. Graph neural networks can handle non-Euclidean space data (graph data) \cite{wu2020comprehensive}, capturing complex interactions between nodes and obtain high-level vector representations of nodes and graphs. Therefore, it can be used to extract spatial relationships between different buildings. Sana Arastehfar et al. \cite{arastehfar2022short} proposed that similar consumption patterns might exist among different households in residential buildings, resulting in analogous electrical loads. Capturing this spatial information among residential units could enhance data-driven load prediction methods. Therefore, they utilized an LSTM model embedded with a GCN for household-level load forecasting. Similarly, Lin et al. \cite{lin2021spatial} considered the influence of similar consumption behavior among residential buildings on load prediction and proposed a load prediction framework based on Graph WaveNet \cite{wu2019graph}. Based on the above two studies, Wang et al. \cite{wang2023short} suggested that the correlation of consumption patterns among households is affected by a variety of complex and unknown factors that may be both naturally linear and nonlinear. They introduced a short-term household load prediction method based on multiple correlation temporal graph neural networks (RLF-MGNN) to model multiple inter-dependence relationships among households. However, the load prediction method based on the correlation of consumption behavior between households in the above research is only applicable to residential buildings and is difficult to be widely promoted as a universal load prediction method in the field of energy load prediction. Furthermore, building energy systems have complex topology, strong non-linearity, time-lag effect, etc. And building energy consumption as typical time series data is not only related to time but is easily influenced by external conditions such as outdoor weather, the physical characteristics of the building, the energy needs of the users, geographical factors, etc. Therefore, it is limited to simply use historical load data and inter-user consumption relationships for modeling.

To address the above-mentioned challenges, our research proposes a novel method of multi-building load prediction, utilizing spatio-temporal graph neural networks \cite{yu2017spatio}. Specifically, the major contributions are three folds:

Firstly, a graph representation method for inter-building correlation is proposed. This method leverages the inherent characteristics of buildings and the external environmental data to identify the operational conditions of each building. Based on this, the building similarity index describing the similarity of energy consumption patterns between buildings is proposed. The graph is then constructed by using the index. An adaptive filtering mechanism is employed in this model to dynamically adjust the graph structure, thereby precisely capturing spatial correlations among buildings. This method offers a general framework for graph-based representation of inter-building correlations in the energy field. 

Secondly, based on this, a multi-level spatio-temporal graph convolution architecture based on the attention mechanism for predicting the energy load of multiple buildings is designed. This method significantly improves the accuracy of building load prediction. 

Finally, a model interpretability method based on the optimal graph structure obtained during the training process is developed. This method reveals the clustering results of building nodes in the feature space, providing new insights for experts to understand the model's principles and evaluate its reliability.

Rest of the paper is structured as follows. In section 2, several related models are introduced. Section 3 elaborates methodology that includes the overview of the model, data preprocessing, Graphical description of relationship among buildings and spatio-temporal dependency extraction. Section 4 provides results simulation and discussion while Section 5 concludes the paper.
\begin{figure*}[t!]
	\centering
	\includegraphics[width=\textwidth]{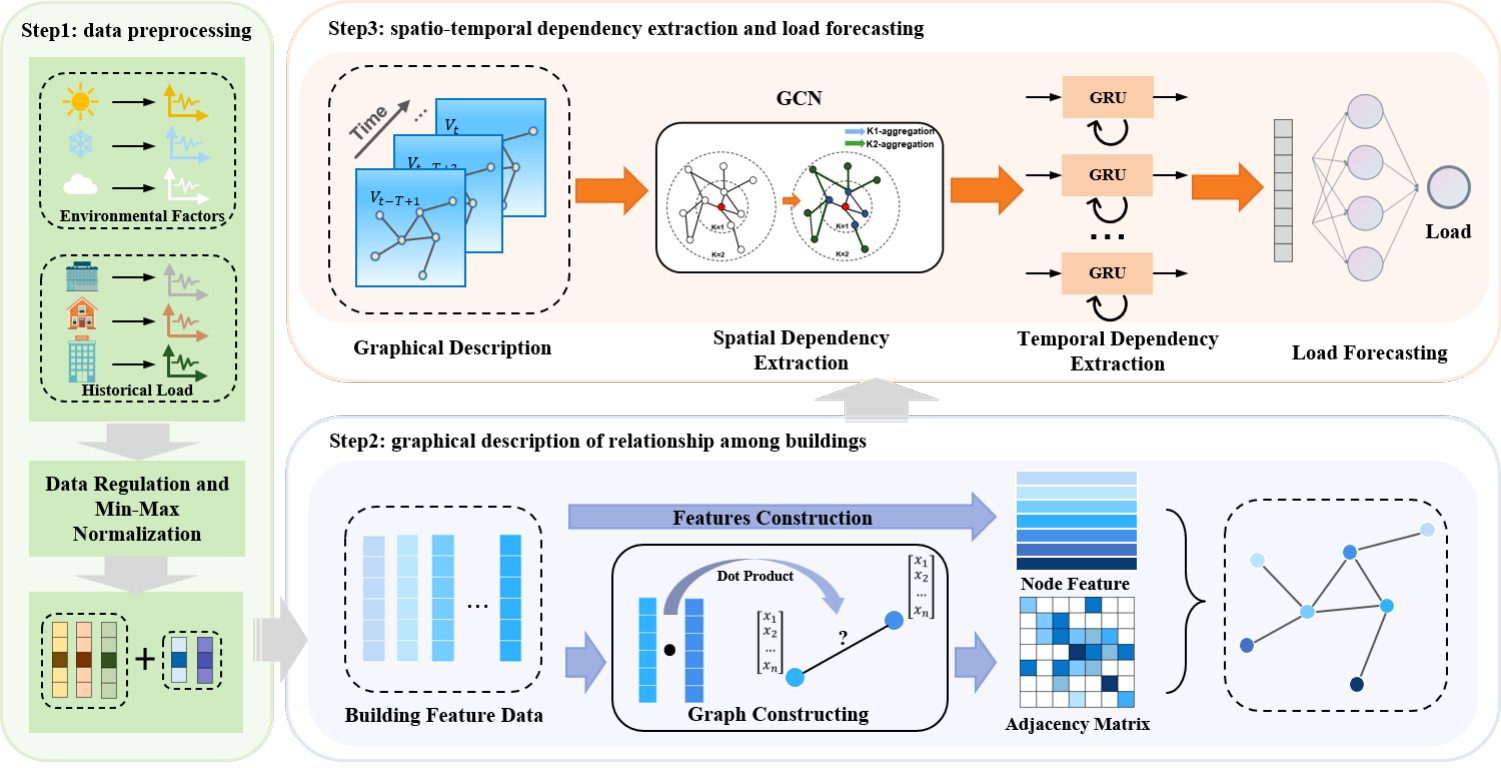} %
	\caption{Schematic of the multi-buildings load forecasting method.}
	\label{table 1}
\end{figure*}

%% file: 02_related_work.tex
\section{Related Methods}
\subsection{Spatio-Temporal Graph Convolutional Networks}
In recent years, there has been a growing trend in research efforts to combine GCN with various temporal processing methods to extract both temporal and spatial information from datasets in the fields of transportation, energy, environment, etc. \cite{jin2023spatio} This hybrid neural network is commonly known as Spatio-Temporal Graph Neural Network (STGNN). Such methods primarily consist of two fundamental modules: a spatial dependency learning module and a temporal dependency learning module. The STGNN used in this paper mainly adopts two core methods: the graph convolution module and a Gated Recurrent Unit (GRU).

\subsubsection{Graph Convolutional Network}
The core structure of graph convolutional neural networks \cite{kipf2016semi} typically comprises convolutional and pooling layers. For node-level tasks, stacking multiple convolutional layers to obtain high-level node representations is often sufficient without the need for pooling layers. However, for graph-level tasks, it is necessary to add pooling layers to generate a global representation of the entire graph, typically following the final convolutional layer. Convolutional layers aggregate information from target nodes and their neighboring nodes to update the representation of the target node using aggregation functions, as shown in the Eq.(1). Pooling layers then generate graph-level representations based on node features and the graph structure. Common pooling methods include global pooling \cite{gilmer2017neural}, virtual node pooling \cite{pham2017graph}, Top K pooling and hierarchical pooling\cite{ying2018hierarchical}\cite{ma2019graph}.
\begin{equation}
    H^{(l+1)}=\sigma(\tilde{D}^{-\frac{1}{2} } \tilde{A}\tilde{D}^{-\frac{1}{2} } H^{(l)} W^{(l)})
\end{equation}

\noindent where \( \bold{\tilde{A}=A+I}\), \(\bold{\quad A \in \mathbb{R}^{n \times n}} \) is the adjacency matrix, \(\bold{\quad I \in \mathbb{R}^{n \times n}}\) is the identity matrix, \(\bold{\tilde{D} \in \mathbb{R}^{n \times n}}\) is the degree matrix. \(\bold{\sigma(\cdot)} \) is the activation function to improve the capacity of nonlinear expressiveness. \(\bold{H^{(l)}}\) is the representation matrix in the lth layer, and \(\bold{W^{(l)}}\) is the trainable weight matrix. For the first layer, \(\bold{H^{(0)} = X}\), \(\bold{\quad X \in \mathbb{R}^{n \times r}}\) is the feature matrix, where \(\bold{r}\) is the dimension of the node feature vector.

\subsubsection{Gated Recurrent Unit (GRU)}
In recent developments within deep learning research, Recurrent Neural Networks (RNNs) \cite{elman1990finding} have become a key instrument for processing sequential data. Nevertheless, a significant challenge for traditional RNNs is their struggle with long-term dependencies, which impedes their ability to capture important information in long sequences. To address this obstacle, a range of improved RNN models have been proposed, among which the Gated Recurrent Unit (GRU) stands out. Compared to the Long Short-Term Memory (LSTM) \cite{hochreiter1997long} network—another variant of RNN—the GRU \cite{cho2014learning} features a simpler structure with fewer parameters. And it’s also capable of effectively capturing long-term dependencies within sequences, exhibiting superior performance in many sequence modeling tasks. The structure of the GRU is illustrated in the Fig.A1(a) in the appendix. The unit at current moment receives the state vector from the previous cell and the feature vector from the current cell. Subsequently, update and reset gates are computed using specific formulas, with values ranging from 0 to 1. The update gate plays a pivotal role in determining the degree to which the state information from the prior moment influences the current state. A higher value of the update gate signifies a greater incorporation of past state information. The reset gate determines the amount of previous state information that is integrated into the current candidate set \(\bold{\tilde{h}_{t}}\), facilitating a nuanced balance between maintaining historical data and new information.
\begin{equation}
    z_{t}=\sigma \left (W_{z}\cdot \left [  h_{t-1},x_{t}\right ]\right ) 
\end{equation}
\begin{equation}
    r_{t}=\sigma \left (W_{r}\cdot \left [  h_{t-1},x_{t}\right ]\right ) 
\end{equation}

\noindent where \(\bold{\tilde{h}_{t}}\) denotes an intermediary state of the current unit, as illustrated by the equation. This state integrates the input data \(\bold{{x}_{t}}\) and incorporates the outcome of the previous state's processing into the current hidden state., through this method to remember the state of the current moment.

The symbol \(\bold{\tilde{h}_{t}}\) denotes an intermediary state of the current cell, as illustrated by the Eq. (4). The \(\bold{\tilde{h}_{t}}\) integrates the input data \(\bold{{x}_{t}}\) and incorporates the outcome of the previous state's processing into the current hidden state. This approach is employed to encode the information pertinent to the current moment.
\begin{equation}
    \tilde{h}_{t}=tanh \left (W\cdot \left [r_{t}\ast h_{t-1},x_{t}\right ]\right ) 
\end{equation}

The concluding phase involves updating the memory content, where the model simultaneously executes the reset and update processes. This is achieved through the update gate \(\bold{{z}_{t}}\), which modulates the extent of information retention and the incorporation of new information into the current unit.
\begin{equation}
    h_{t}=\left (1-z_{t}\right ) \ast h_{t-1}+z_{t} \ast \tilde {h}_{t} 
\end{equation}
\subsection{Attention Mechanism}
As illustrated in the Fig. A1(b) in the appendix, the attention mechanism \cite{vaswani2017attention} compresses all significant information from a vector sequence into a single context vector \(\bold{c}\). In detail, it evaluates the relevance of each sequence element to others by computing similarity scores, which are then normalized to form attention weights. These weights are used to generate the attention mechanism's output through a weighted summation with corresponding element, effectively focusing on the most relevant information. 

Initially, the input vectors \(\bold{{x}_{i}}\) undergo transformation through multiplication with trainable matrices \(\bold{W^{q}}\), \(\bold{W^{k}}\), \(\bold{W^{v}}\), producing the query (\(\bold{{q}_{i}}\)), key \(\bold{({k}_{i}}\)), and value (\(\bold{{v}_{i}}\)) vectors, respectively.
\begin{equation}
    q_{i}=W^qx_{i}
\end{equation}
\begin{equation}
    k_{i}=W^kx_{i}
\end{equation}
\begin{equation}
    v_{i}=W^vx_{i}
\end{equation}

Subsequently, the procedure calculates the similarity between two vectors utilizing the Eq. (9), and the similarity is normalized through the softmax function to ensure the similarities sum to one, where {\(\bold{d_{k}}\)} denotes the dimension of the key vectors.
\begin{equation}
    \alpha_{1,i}=\frac{q_i\cdot k_i}{\sqrt{d_k}}
\end{equation}
\begin{equation}
    {\alpha}' _{1,i}=\frac{e^{\alpha_{1,i}}}{\sum _{j} e^{\alpha_{1,j} } } 
\end{equation}

Finally, the attention-enhanced output vector is generated through a weighted summation operation according to the Eq. (11)
\begin{equation}
    x'_1={\sum}_i\alpha'_{1,i} v_i
\end{equation}

Viewing these operations through the perspective of matrix algebra, the entire methodology can be briefly expressed by the following Eq. (12).
\begin{equation}
    Attention\left(Q,K,V\right) = Softmax\left(\frac{QK^T}{\sqrt{d_k}}\right)V
\end{equation}
\begin{figure*}[t!]
	\centering
	\includegraphics[width=\textwidth]{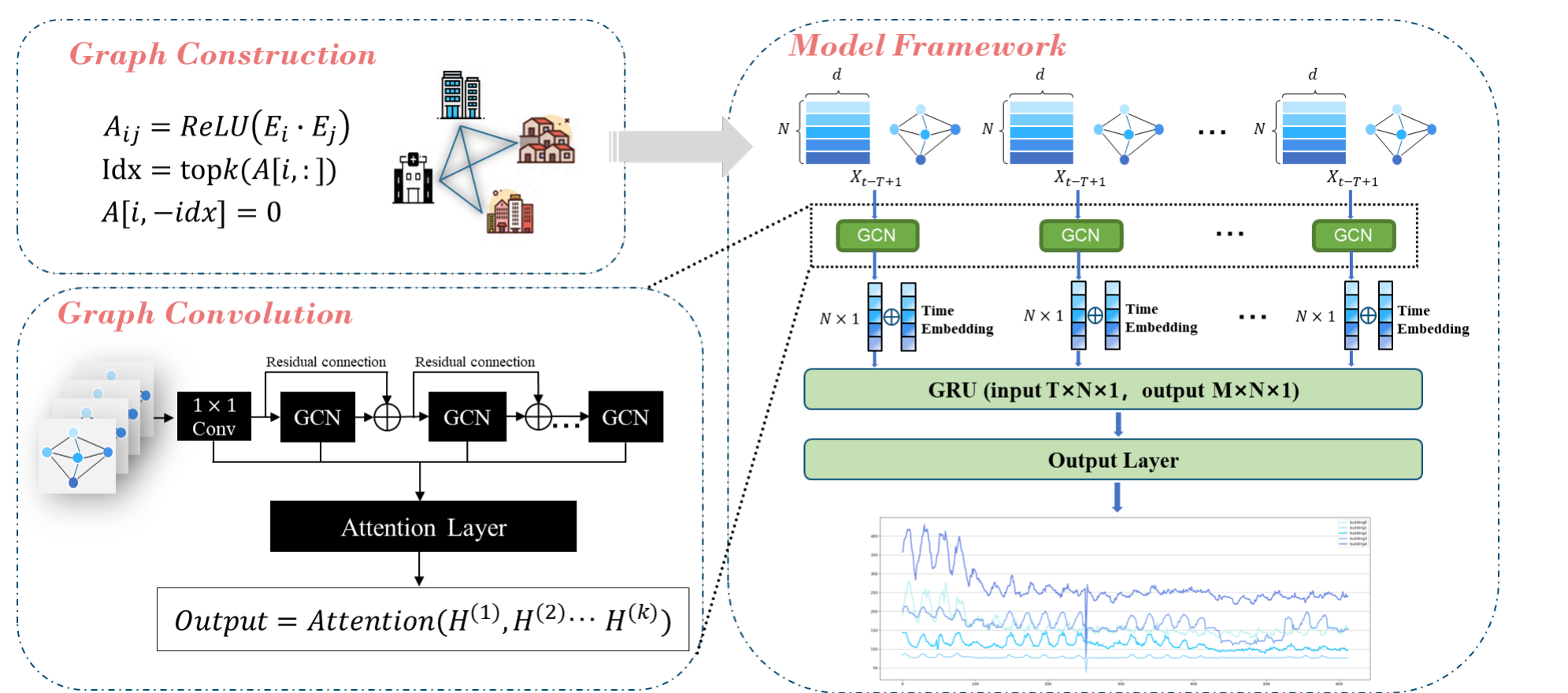} %
	\caption{Framework of the proposed model.}
	\label{table 1}
\end{figure*}

%% file: 03_methodology.tex
\section{Methodology}
The flow chart of the multi-building load forecasting method based on STGNN (Spatio-Temporal Graph Neural Network)is shown in the Fig. 1. Firstly, the historical building load data and external environmental data are formatted and normalized using min-max normalization to obtain the building feature vectors that represent various building energy consumption patterns. Secondly, a method for constructing the graph structure based on the similarity of energy consumption patterns between buildings is proposed. Finally, the spatio-temporal convolution module extracts both the spatial dependencies of buildings and the temporal dependencies within historical load data from the multi-building dataset, then conducts load forecasting. We will begin with a broad overview of the model structure, followed by a detailed explanation of the above-mentioned three stage.

\subsection{Overview}
The structure of the model is shown in the Fig. 2. It is composed of a graph constructing module, a graph convolution module, a GRU module, and a load prediction module. This model is designed for short-term load forecasting, which aims to predict future load for the next few steps based on historical building load \cite{gross2005short}. Let \(\bold{X=\left \{ x_{t1},x_{t2},\cdots ,x_{tT}  \right \}} \) denote the historical load data of the building, where \(\bold{x_{ti}}\) is the energy load at a specific historical time point \(\bold{i}\), and \(\bold{T}\) represents the total number of historical time steps. \(\bold{Y=\left \{ y_{tT+1},y_{tT+2},\cdots ,x_{tT+M}  \right \}} \) represents the future \(\bold{M}\) steps of building load to be predicted. The type of data input into the model in this study is the spatio-temporal graph. The spatio-temporal graph data extends traditional graph data with an additional time dimension. Therefore, the model input is \(\bold{X\in \mathbb{R}^{T\times N\times d} }\), where \(\bold{T}\) denotes the number of time steps of historical data, \(\bold{N}\) is the number of nodes in the graph (i.e., the number of buildings), and \(\bold{d}\) represents the dimension of the features of each node in the graph. Another input to the model is the adjacency matrix \(\bold{A\in \mathbb{R}^{N\times N}}\), which represents the connection relationship in the graph and is produced by the Graph Constructing module. The magnitude of the element in the adjacency matrix represents the correlation between the two nodes in the graph. The adjacency matrix and historical load data are initially processed by the Graph Convolution module to extract spatial correlations from the historical data, as depicted by Eq. (13).
\begin{equation}
    X\in \mathbb{R}^{T\times N\times d},A \xrightarrow{Graph\ Convolution} X'\in \mathbb{R}^{T\times N\times 1}
\end{equation}

Following graph convolution, the output is sent to the temporal learning module, which is responsible for extracting the temporal information hidden in the data.
\begin{equation}
    X'\in \mathbb{R}^{T\times N\times 1},A \xrightarrow{Temporal\ Learning} \hat{Y}\in \mathbb{R}^{M\times N\times 1}
\end{equation}

In order to generate the final output, the load forecasting module processes the data, which has fully explored the spatio-temporal dependencies of the historical data, and maps it to the required output dimensions, which is shown in Eq. (14).

\subsection{Data preprocessing}
To identify different energy consumption patterns of buildings, we select certain features that effectively represent the operating conditions of these buildings. These features include historical load data, temporal information, and external environmental data, etc. These data are normalized to the range of 0 to 1 by the min-max normalization, which can eliminate the impact of different dimensions of features on load prediction. Subsequently, the normalized data is transformed into a suitable format to establish the feature vector \(\bold{x}\) for each individual node, thereby constructing the feature matrix \(\bold{X}\) for the entire graph.

\subsection{Graphical description of relationship among buildings}
Based on the characteristics of the buildings themselves and external environmental information, each building is mapped to a feature space, where each building is considered a node in the graph. In this vector space, each building is represented by a feature vector, with buildings exhibiting similar energy consumption patterns having highly similar feature vectors. Consequently, the similarity between feature vectors can be leveraged to characterize the spatial dependence among buildings and to establish the relationship graph among them. We define the building similarity index \(\bold{a}\) to describe the similarity between buildings as shown in Eq. (15), where \(\bold{E}\) represents the building's feature vector determined by the building's inherent characteristics and external environmental variables. \(\bold{Similarity\left(\cdot\right)}\) is the building similarity calculation formula, which varies depending on the task.
\begin{equation}
    a=Similarity\left(E_i\cdot E_j \right)
\end{equation}

In this study, cosine similarity \cite{tan2016introduction} is used as similarity calculation formula to express this correlation. Cosine similarity is a metric of the degree of similarity in the direction of two vectors. It determines the similarity by calculating the ratio of the dot product of two vectors to the product of their modulus, with values ranging from -1 (completely opposite) to 1 (completely the same). In order to prevent buildings with significantly different energy consumption patterns from affecting the target building and to minimize computational complexity during model learning, this study introduces an edge filtering mechanism.    In the adjacency matrix of building similarity, each row represents the similarity between the target building and other buildings. The top k elements with the highest similarity values are retained, while the remaining elements are set to zero. Before the model training, the initialized adjacency matrix is constructed based on the load data and outdoor environmental data for the whole year, and then the adjacency matrix will continuously adjust locally based on the current input data during model training. This approach not only considers the global historical information but utilizes the local detail information. It facilitates the identification of building energy consumption patterns and the extraction of spatial relationships among buildings at multiple scales. In general, the graphical representation of inter-building correlations is established according to the Eq. (15)-(17). The correlation between the nodes is calculated based on the Eq. (15). The Eq. (16)-(17) illustrates the edge filtering mechanism. Throughout the training process, the adjacency matrix is adaptable to change with the training data and updates the model parameters, so that the structure of building correlation graph can reach the optimal state.
\begin{equation}
A_{i,j}=ReLU\left(E_i\cdot E_j\right)
\end{equation}
\begin{equation}
idx=topk\left(A\left[i,:\right]\right)
\end{equation}
\begin{equation}
A\left[i,-idx \right]=0
\end{equation}

\noindent where \(\bold{A_{i,j}}\) represents the element in the \(\bold{i}\)th row and \(\bold{j}\)th column of the adjacency matrix, and the function \(\bold{topk\left(\cdot\right)}\) outputs the indices of the top \(\bold{k}\) values in the ith row of the adjacency matrix.
\subsection{Spatio-temporal dependency extraction and load forecasting}
After the graph constructing module, we have generated a graphical depiction of the relationships among various buildings. This phase involves extracting the spatial dependencies among buildings and the temporal dependencies from the historical load, and then make load predictions. 

The extraction of spatial dependencies aims to fuse the information of a given node with its neighboring nodes, thereby handling the spatial dependencies in the graph. We will first provide a detailed mechanism of this procedure in Eq. (19), followed by an explanation of our motivation.
\begin{equation}
    H_{out}=Attention\left(H^{\left(1\right)},H^{\left(2\right)},\cdots,H^{\left(k\right)}\right)
\end{equation}

\noindent where \(\bold{K}\) represents the depth of propagation (i.e., the layers of graph convolution), \(\bold{H^{(i)}}\) represents the hidden state of the \(\bold{i}\)th layer, \(\bold{H_{out}}\) represents the output hidden state of the graph convolution network module, and \(\bold{Attention(\cdot)}\) represents the attention mechanism. The graph convolution part in Fig. 2 demonstrates the process of information propagation based on graph convolution horizontally and the aggregation process of hidden states based on the attention mechanism vertically. This method is designed to address the issue of over-smoothing \cite{li2018deeper} in GCNs. In GCNs, after multiple layers are incorporated, the features of different nodes tend to homogenize, irrespective of the initial state of the nodes. Conversely, insufficient layers impede effective information propagation and the extraction of high-dimensional spatial features. Therefore, the attention mechanism is incorporated into this network to facilitate deeper convolution operations while preserving node-specific differences. Similarly, to overcome the over-smoothing problem, and to prevent gradient vanishing and network non-convergence, residual connections \cite{he2016deep} are added into the graph convolution network.

The output from the spatial dependency extraction module contains the spatial dependency relationships among different buildings. Given that building load is time-series data, the model should be capable of extracting temporal dependencies across various time steps. Consequently, in the Temporal Dependency Extraction phase, the GRU model is added to capture the temporal dependencies of each node. As depicted in the Fig. 2, a sequence of time series is fed into multiple GRU units horizontally, and multiple layers can be stacked vertically to further extract time information.

From the preceding GRU module, we obtained the node-level feature representation. Each node's representation already incorporates the temporal and spatial dependencies in the historical load data of buildings. So we simply need to map each node's feature representation to the sample label space to predict the future load values of each building. The fully connected layer is adopted to learn the complex relationship between the global representation and the load labels.

Finally, the Mean Squared Error (MSE) loss function is employed, as defined in the Eq. (20).  The model is trained by minimizing the error between the load labels and the predicted values in the training set.
\begin{equation}
    loss=\frac{1}{n}\sum_{i=1}^{n}\left(y_i-\hat{y}_i\right)^2
\end{equation}

\noindent where \(\bold{n}\) represents the number of data samples in the training set, \(\bold{y_{i}}\) represents the real value of the \(\bold{i}\)th sample, and \(\bold{\hat{y} _{i} }\) represents the predicted value of the \(\bold{i}\)th sample.

\subsection{Model interpretability}

To enhance the practicality of the model in real-world scenarios, the model needs to be interpreted after making predictions to explain why it is effective. As mentioned earlier, the proposed model adaptively generates and modifies the building correlation graph structure during the training phase. The final graph structure after training will be visualized to show the connectivity between different building nodes. Then, based on the connectivity relationships, the building nodes are classified, and the classification results are compared with the K-means clustering outcomes. Finally, to further explain the rationality of the clustering results, the load curves of buildings in each cluster are visualized to compare the differences in load levels between clusters and the similarities of load curves within the same cluster.

%% file: 04_results.tex
\section{Results}
\begin{table}[h]
    \centering
    \begin{tabular}{lll}
        \toprule
        Method & Hyper-parameters & Optimal values \\
        \midrule
        MTGNN & Number of hidden layers & 4 \\
        FCNN & Number of hidden layers & 4 \\
             & Number of neurons in hidden layer & 128 \\
        SVR & C & 1.7 \\
            & gamma & 0.6 \\
        XGBoost & Max depth & 3 \\
                & Learning Rate & 0.4 \\
                & Number of estimators & 100 \\
                & Subsample & 0.8 \\
        GRU & Number of hidden layers & 2 \\
        \bottomrule
    \end{tabular}
    \caption{The optimal hyperparameters of MTGNN, FCNN, SVR, XGBoost, and GRU.}
    \label{tab:optimal_hyperparams} 
\end{table}

\begin{table}[h]
    \centering
    \scalebox{0.82}{
    \begin{tabular}{lll}
        \toprule
        Parameter & Value & Description \\
        \midrule
        GCN\_conv\_channels & 16 & Dimension of each GCN convolution layer \\
        GCN\_hidden\_layers & 4 & Number of stacked GCN blocks \\
        GCN\_depth & 2 & Number of each GCN hidden layers \\
        GRU\_layer & 2 & Number of GRU hidden layers \\
        GRU\_dim & 16 & Dimension of GRU layers \\
        att\_dim & 32 & Dimension of attention mechanism \\
        acti\_f & ReLU & Activation function used in model layers \\
        dropout\_p & 0.3 & Dropout probability to prevent overfitting \\
        lr & 0.001 & Learning rate \\
        loss & MSE & Loss function used to train the model \\
        batch\_size & 32 & Training batch size \\
        train\_epoch & 100 & Number of training epochs \\
        \bottomrule
    \end{tabular}}
    \caption{Model Architecture Parameters.}
    \label{tab:model_params} 
\end{table}

\subsection{Dataset}
The proposed method is evaluated on the Building Data Genome Project 2 dataset. The dataset contains hourly sensor data over the full breadth of 2016 to 2017 from 1636 buildings across 19 different regions worldwide. The data includes Electrical load, Heating load, Cooling load, Steam, Solar energy, Water, and Irrigation; meteorological data includes outdoor temperature, relative humidity, dew point temperature, barometric pressure, wind speed, wind direction, and cloud coverage. A detailed dataset introduction is available in the reference \cite{miller2020building}. After the data cleaning and data preprocessing, this study generated two subsets, a larger one and a smaller one, which contain hourly operational data for 20 buildings and 500 buildings in 2016, respectively, with a total of 8784 time points. In this study, our experiments primarily focus on building electricity load forecasting. Building electricity load is significantly affected by two factors: building occupancy and outdoor conditions. Since the occupancy schedule of buildings with specific functions is usually fixed and related to time, time variables can be used to consider the impact of building occupancy. Outdoor conditions can be well described using variables such as outdoor temperature, dew point temperature, outdoor relative humidity, wind direction, and wind speed. Hence, the dataset's feature set comprises these environmental parameters alongside time variables (month, day, hour, minute, and day type), which have a significant impact on the buildings' load profiles. The data for the whole year has been normalized using min-max normalization to mitigate the impact of disparate scales on results, with the normalization Eq. (21) as follows. The data is divided into training, testing, and validation sets in a time-sequential order at a ratio of 8:1:1.
\begin{equation}
    \hat{x}=\frac{x-min\left(X\right)}{max\left(X\right)-min\left(X\right)}
\end{equation}

\noindent where \(\bold{x}\) represents the original data, \(\bold{\hat{x}}\) signifies the normalized data, and \(\bold{max(X)}\) and \(\bold{min(X)}\) respectively represent the maximum and minimum values in the dataset.
\subsection{Results comparison}
\subsubsection{Experimental setup}
\begin{figure*}[t!]
    \centering
    \subfloat[]{ 
        \includegraphics[width=6.5cm]{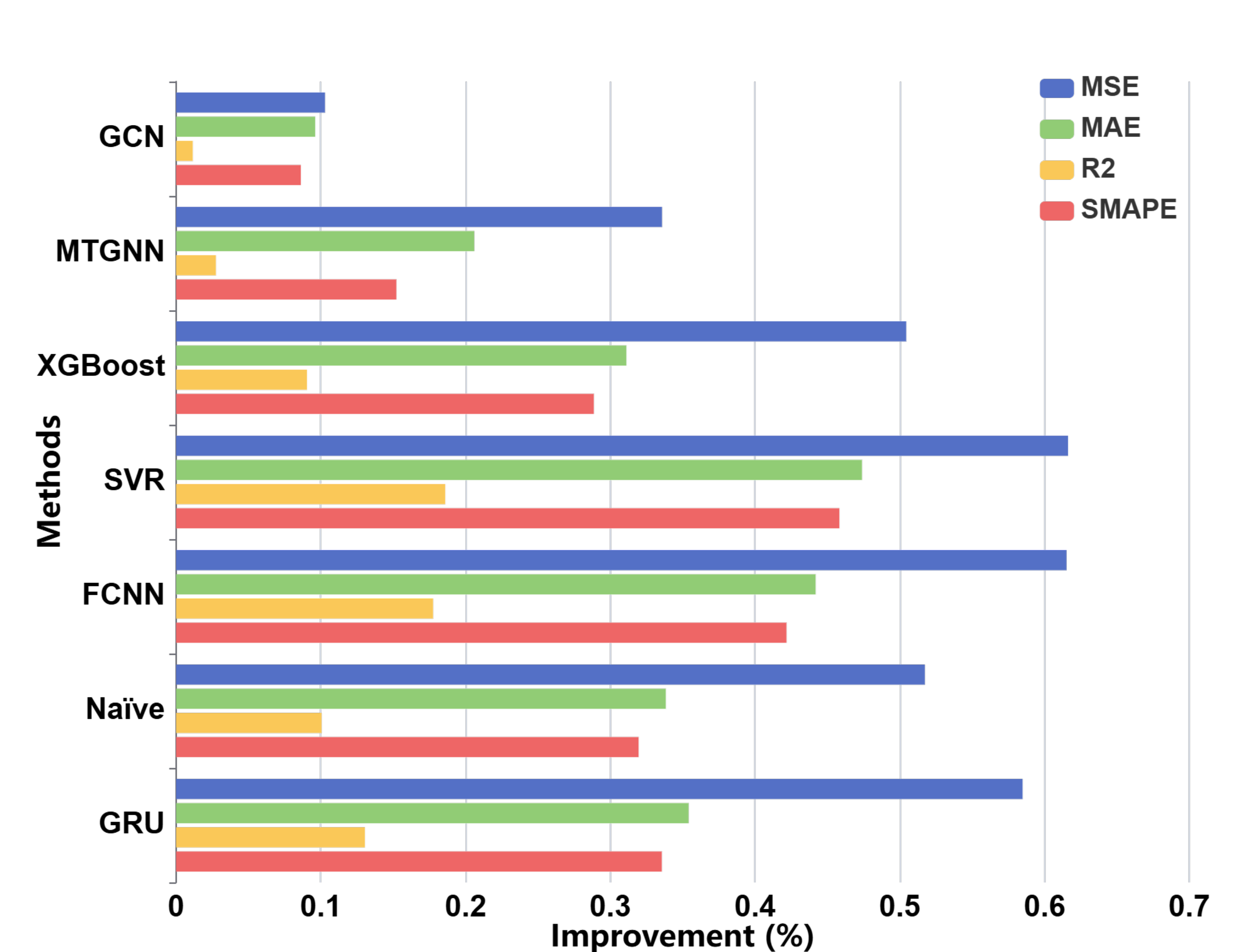} 
        \label{Fig.sub.1}} 
    \subfloat[]{ 
        \includegraphics[width=6.5cm]{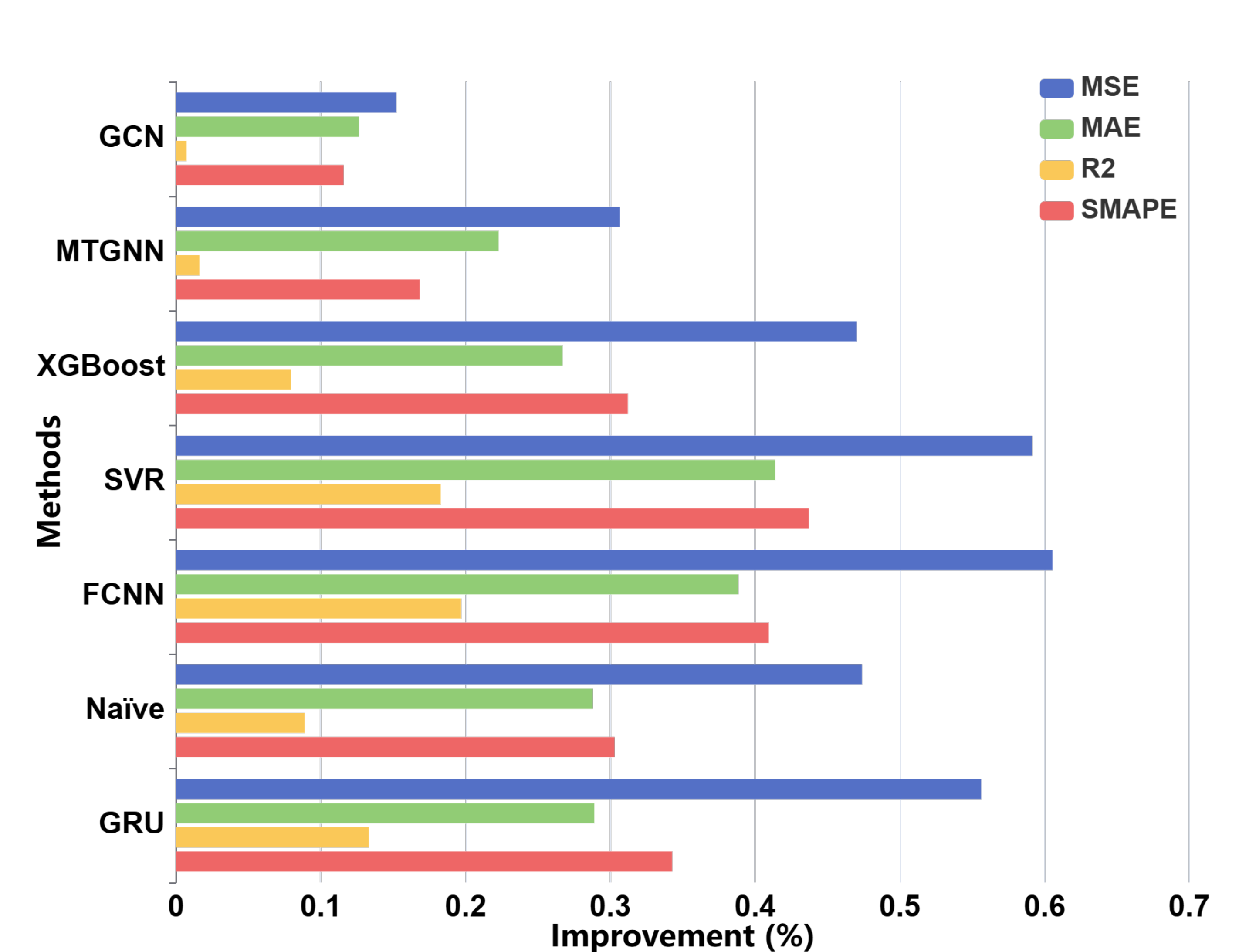} 
        \label{Fig.sub.2}} 
    \caption{The improvement percentage from various models with Att-GCN.(a) The improvement percentage with 20 buildings.(b) The improvement percentage with 500 buildings.}
\end{figure*}
To evaluate the performance of the proposed model that considers the spatio-temporal relationships among buildings, XGBoost, SVR, FCNN, GRU, and Naïve are selected as baseline models for accuracy comparison with the proposed model. XGBoost \cite{chen2016xgboost} is an efficient and flexible tree-based machine learning algorithm, particularly suitable for regression and classification problems, which improves prediction performance by integrating multiple weaker learners. Support Vector Regression (SVR) \cite{platt1998sequential} is a machine learning algorithm based on Support Vector Machines (SVM) \cite{drucker1996support} for solving regression problems, predicting by finding the optimal hyperplane in high-dimensional space.  Fully Connected Neural Network (FCNN) is a neural network architecture where each neuron in a layer is connected to every neuron in the next layer, enabling comprehensive feature learning and complex nonlinear function approximation. GRU is a variant of recurrent neural network structure with update gates and reset gates to effectively capture long-term dependencies within sequential data, making it particularly adept for time series analysis and related tasks. The Naïve method is a simple baseline model in time series processing, which uses the most recent observation as the future prediction value, suitable for quickly establishing a performance benchmark or evaluating model efficacy. These five classic building load prediction models cannot handle graph-structured data, hence cannot capture the spatial dependencies among buildings. The input features of the first four baseline models include historical building load (the previous 12 hours), environmental factors (outdoor temperature, dew point temperature, wind direction, and wind speed), and time information (day type). Compared with these baseline models, MTGNN \cite{wu2020connectingdotsmultivariatetime} and the proposed Att-GCN and GCN models are capable to capture the complex spatial dependencies among buildings. The input of these three models is the spatio-temporal graph data of the previous 12 hours, with the node feature attributes including the current load value, day type, outdoor temperature, dew point temperature, wind direction, and wind speed. The main difference between proposed Att-GCN and GCN is that Att-GCN adopts an attention mechanism-based hierarchical graph convolution framework in the graph convolution module. Apart from this key difference, the configuration of the two models remains identical. The optimal hyperparameters of all baseline models are obtained through grid search and cross-validation, as shown in the Table \ref{tab:optimal_hyperparams} and the detailed model architecture of our att-GCN can be found in Table \ref{tab:model_params}. Att-GCN and GCN are implemented using PyTorch \cite{pytorch}. The model is trained using the Adam optimizer \cite{kingma2014adam} with a learning rate of 0.001. The batch size is 32, and the epoch number is 100. Model prediction accuracy was evaluated using MSE (Mean Square Error), MAE (Mean Absolute Error) metrics, $R^2$ (Coefficient of determination) and SMAPE (Symmetric Mean Absolute Percentage Error), which is computed as Eq.~(\ref{eq:mse}), ~(\ref{eq:mae}), ~(\ref{eq:r2}) and ~(\ref{eq:smape}) respectively.
\begin{equation}
    \label{eq:mse}
    MSE=\frac{1}{n}\sum_{i=1}^{n}\left(y_i-\hat{y}_i\right)^2
\end{equation}
\begin{equation}
    \label{eq:mae}
    MAE=\frac{1}{n}\sum_{i=1}^{n}\left|y_i-\hat{y}_i\right|
\end{equation}
\begin{equation}
    \label{eq:r2}
    R^2 = 1 - \frac{\sum_{i=1}^{n} (y_i - \hat{y}_i)^2}{\sum_{i=1}^{n} (y_i - \bar{y})^2}
\end{equation}
\begin{equation}
    \label{eq:smape}
    SMAPE=\frac{1}{n} \sum_{i=1}^{n} \frac{|\hat{y}_i - y_i|}{(|y_i| + |\hat{y}_i|)/2}
\end{equation}

\noindent where \(\bold{n}\) represents the number of data samples, \(\bold{y_{i}}\) and \(\bold{\hat{y}_{i}}\) respectively represent the real and predicted values. The smaller the MSE and MAE values are, the more accurate the prediction result is. In addition, in the experiment, the past 12 hours of building historical data is leveraged to predict the building load for the next hour.

\begin{table}[h]
    \centering
    \resizebox{\textwidth}{!}{ 
    \begin{tabular}{lllllllll}
        \toprule
        & \multicolumn{3}{l}{Multi-Building Method} & \multicolumn{5}{l}{Single-Building Method} \\
        \midrule
        Method & Att-GCN & GCN &MTGNN & XGBoost & SVR & FCNN & Naïve & GRU \\
        MSE & \textbf{0.0028} & 0.0031 & 0.0042 & 0.0056 & 0.0073 & 0.0073 & 0.0058 & 0.0067 \\
        MAE & \textbf{0.0331} & 0.0366 & 0.0417 & 0.0481 & 0.0629 & 0.0593 & 0.0500 & 0.0513 \\
        $R^2$ & \textbf{0.9490} & 0.9382 & 0.9236 & 0.8703 & 0.8002 & 0.8059 & 0.8623 & 0.8395 \\
        SMAPE & \textbf{0.0986} & 0.1079 & 0.1163 & 0.1386 & 0.1820 & 0.1705 & 0.1449 & 0.1484 \\
        \bottomrule
    \end{tabular}
    }
    \caption{Prediction accuracy of models with 20 buildings}
    \label{tab:prediction_accuracy_20}
\end{table}

\subsubsection{Forecasting results comparison}
The average prediction accuracy of Att-GCN, GCN, MTGNN, XGBoost, SVR, FCNN, Naïve, and GRU on two datasets is shown in the Table \ref{tab:prediction_accuracy_20}. We first evaluate the prediction accuracy of the model on the small dataset of 20 buildings. The results showed that Att-GCN, part of the Multi-Building Method, outperforms all the baselines significantly, with MSE, MAE, $R^2$ and SMAPE values of 0.0028, 0.0331, 0.9490 and 0.0986 respectively.  
\begin{figure*}[t!]
    \centering
    \subfloat[]{ 
        \includegraphics[width=6.5cm]{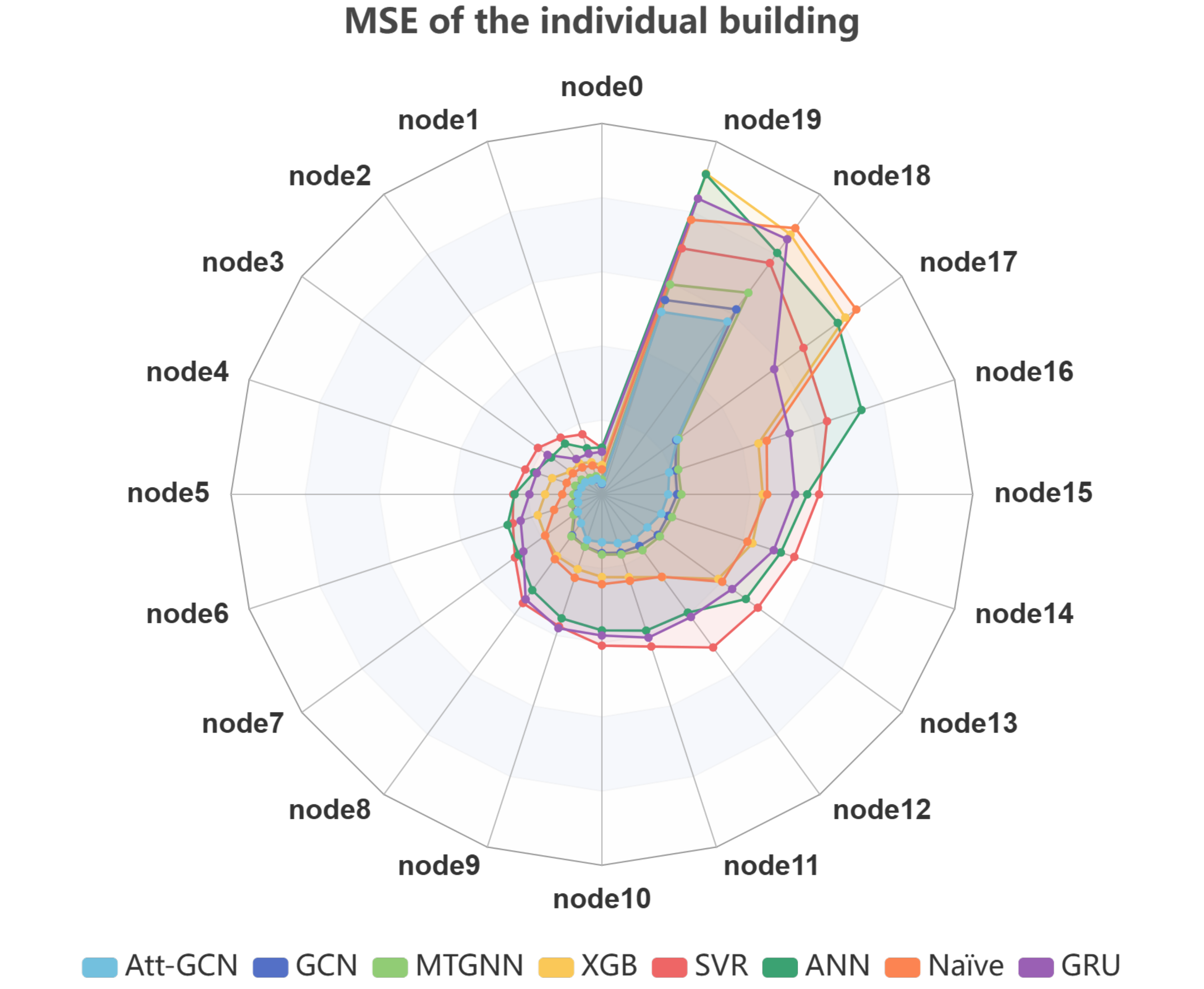} 
        \label{Fig.sub.1}} 
    \subfloat[]{ 
        \includegraphics[width=6.5cm]{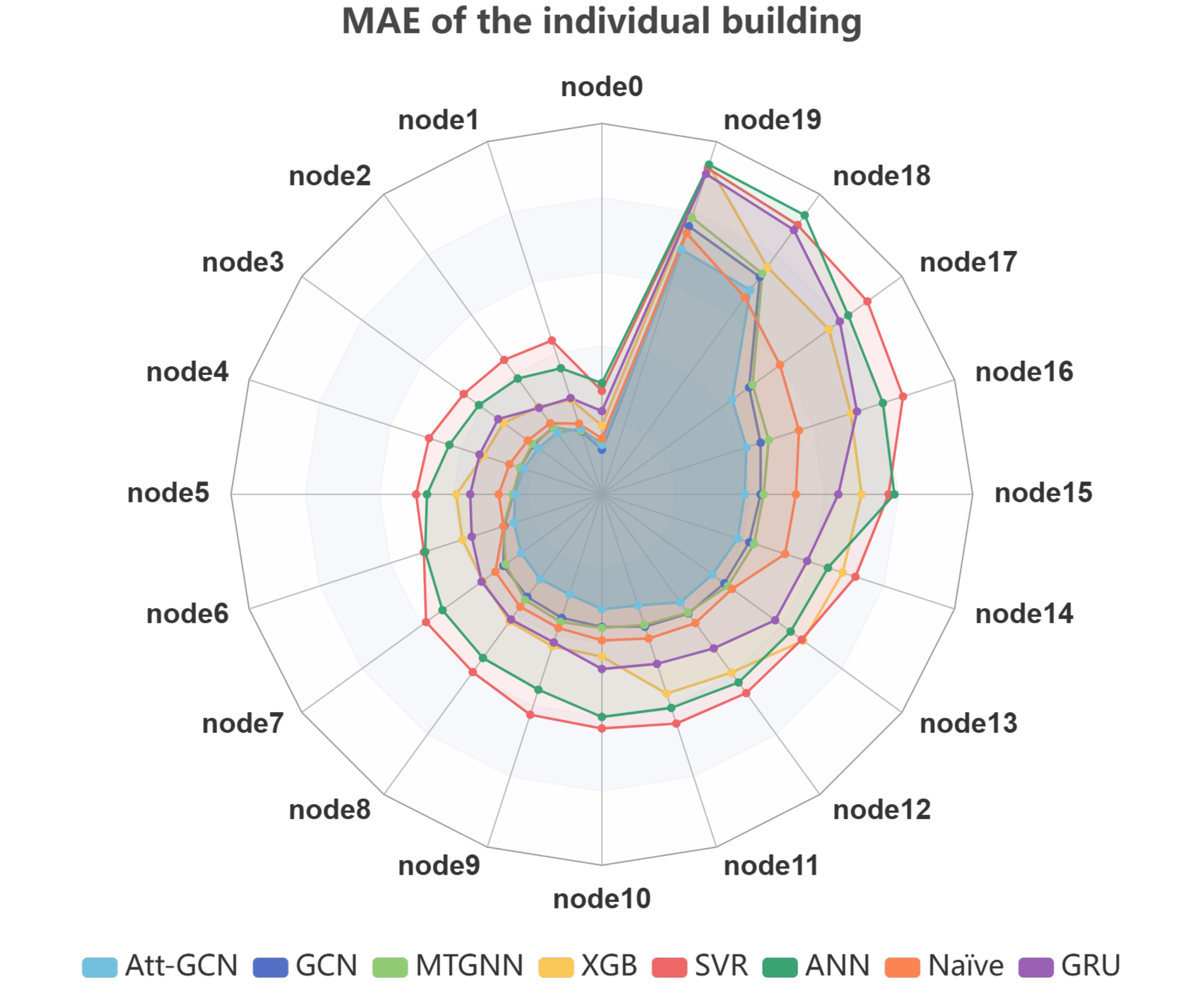} 
        \label{Fig.sub.2}} 
    \caption{Prediction performance of models on the individual buildings. (a) MSE of the individual building. (b)MAE of the individual building.}
\end{figure*}
This significantly outperforms the best baseline model MTGNN, whose MSE, MAE, R² and SMAPE are 0.0042, 0.0417, 0.9236 and 0.1163. As illustrated in the Fig. 3(a), Att-GCN outperforms the MSE of other models by 10.3\%(GCN), 33.5\%(MTGNN), 50.4\%(XGBoost), 61.6\%(SVR), 61.5\%(FCNN), 51.7\%(Naïve) and 58.5\%(GRU). Similarly, the MAE outperformed by 9.6\%(GCN), 20.5\%(MTGNN), 31.1\%(XGBoost), 47.4\%(SVR), 44.2\%(FCNN), 33.8\%(Naïve) and 35.4\%(GRU). In the context of individual buildings, the radar chart Fig. 4 was utilized to visualize the predictive accuracy of the seven methods for each of the 20 buildings. The results indicated that the proposed model consistently outperforms the baseline across all buildings.

\begin{table}[h]
    \centering
    \resizebox{\textwidth}{!}{ 
    \begin{tabular}{lllllllll}
        \toprule
        & \multicolumn{3}{l}{Multi-Building Method} & \multicolumn{5}{l}{Single-Building Method} \\
        \midrule
        Method&Att-GCN& GCN&MTGNN& XGBoost&SVR&FCNN&Naïve&GRU\\
        MSE& \textbf{0.0031} & 0.0037 & 0.0045 & 0.0059 &0.0076 &0.0079 &0.0059 &0.0070\\
        MAE& \textbf{0.0372} & 0.0426 & 0.0479 & 0.0508 &0.0635 &0.0609 &0.0523 &0.0524\\
        $R^2$ & \textbf{0.9285} & 0.9218 & 0.9137 & 0.8601 & 0.7850 & 0.7757 & 0.8527 & 0.8194 \\
        SMAPE & \textbf{0.1047} & 0.1184 & 0.1259 & 0.1522 & 0.1860 & 0.1773 & 0.1502 & 0.1593 \\
        \bottomrule
    \end{tabular}
    }
    \caption{Prediction accuracy of models with 500 buildings}
    \label{tab:prediction_accuracy_500}
\end{table}

Comparative analysis reveals that the Multi-Building Method significantly outperforms the Single-Building Method in forecasting accuracy. Even the relatively inferior models within the Multi-Building category, such as GCN and MTGNN, still surpass the best-performing Single-Building method, XGBoost. The superior performance of the Multi-Building Method can be attributed to its incorporation of spatial dependencies among buildings, enabling the model to identify and learn from the relationships among buildings with analogous energy consumption patterns, which is crucial for accurate load forecasting. In the comparison between Att-GCN and the conventional GCN within the Multi-Building Method, the former exhibits lower MSE, MAE and SMAPE, as well as better $R^2$, suggesting that the integration of attention mechanisms within graph convolutional layers enhances forecasting accuracy. This undoubtedly demonstrates the significant potential of Att-GCN in the field of multi-building load forecasting. 
\begin{figure*}[t!]
    \centering
    \subfloat[]{ 
        \includegraphics[width=6.5cm]{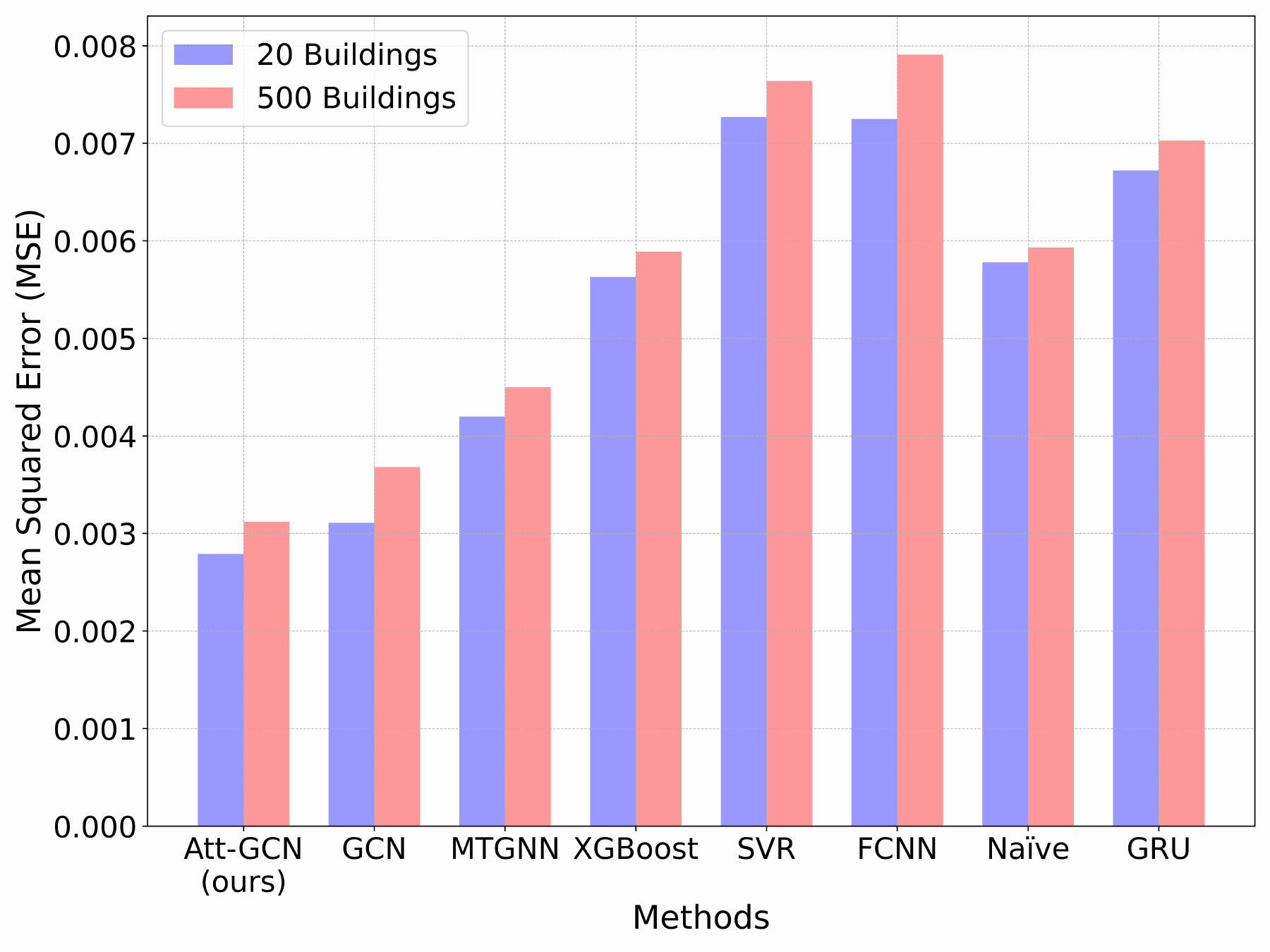} 
        \label{Fig.sub.1}} 
    \subfloat[]{ 
        \includegraphics[width=6.5cm]{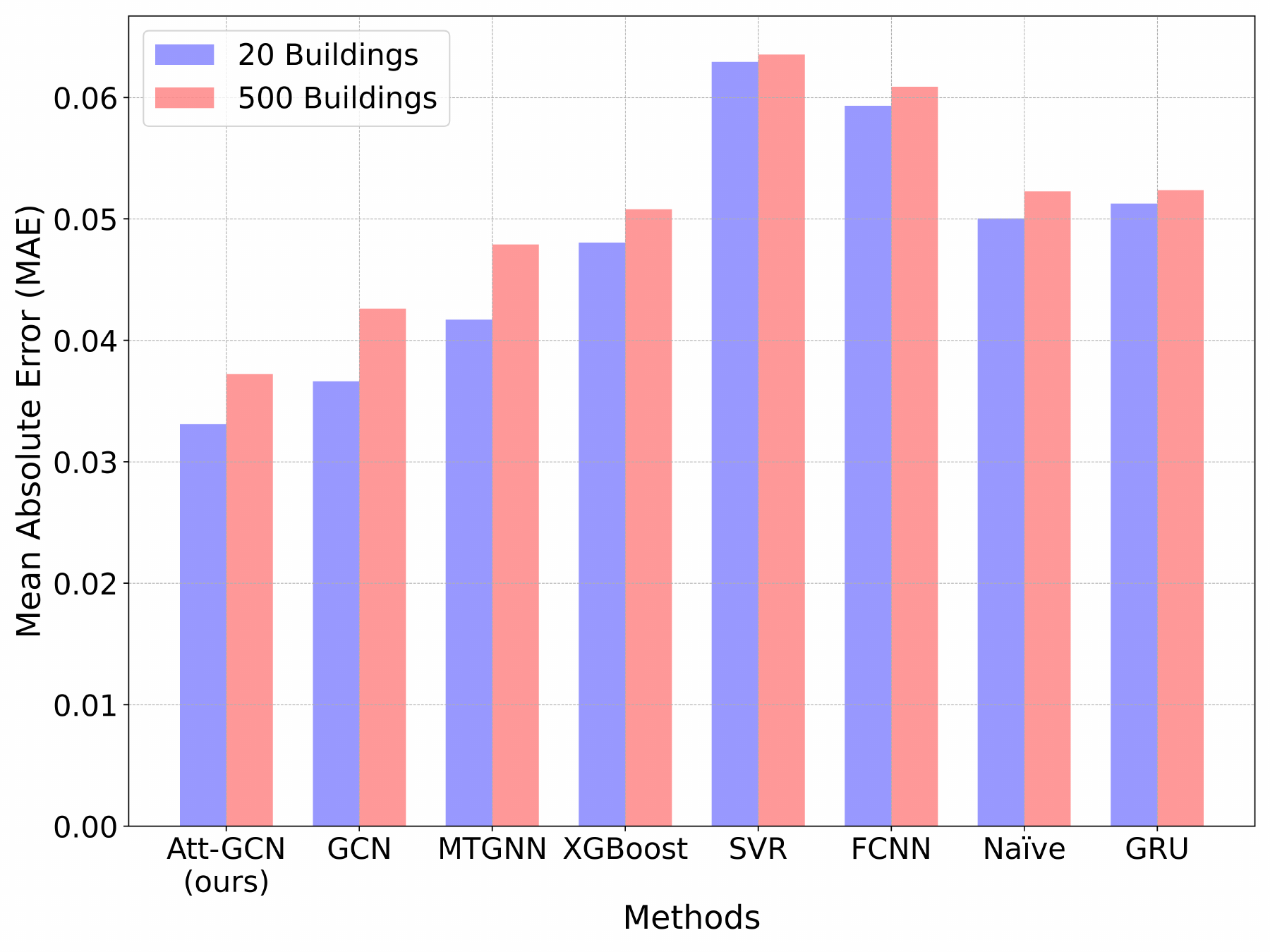} 
        \label{Fig.sub.2}}
    \vspace{-0.3cm}
    \subfloat[]{ 
        \includegraphics[width=6.5cm]{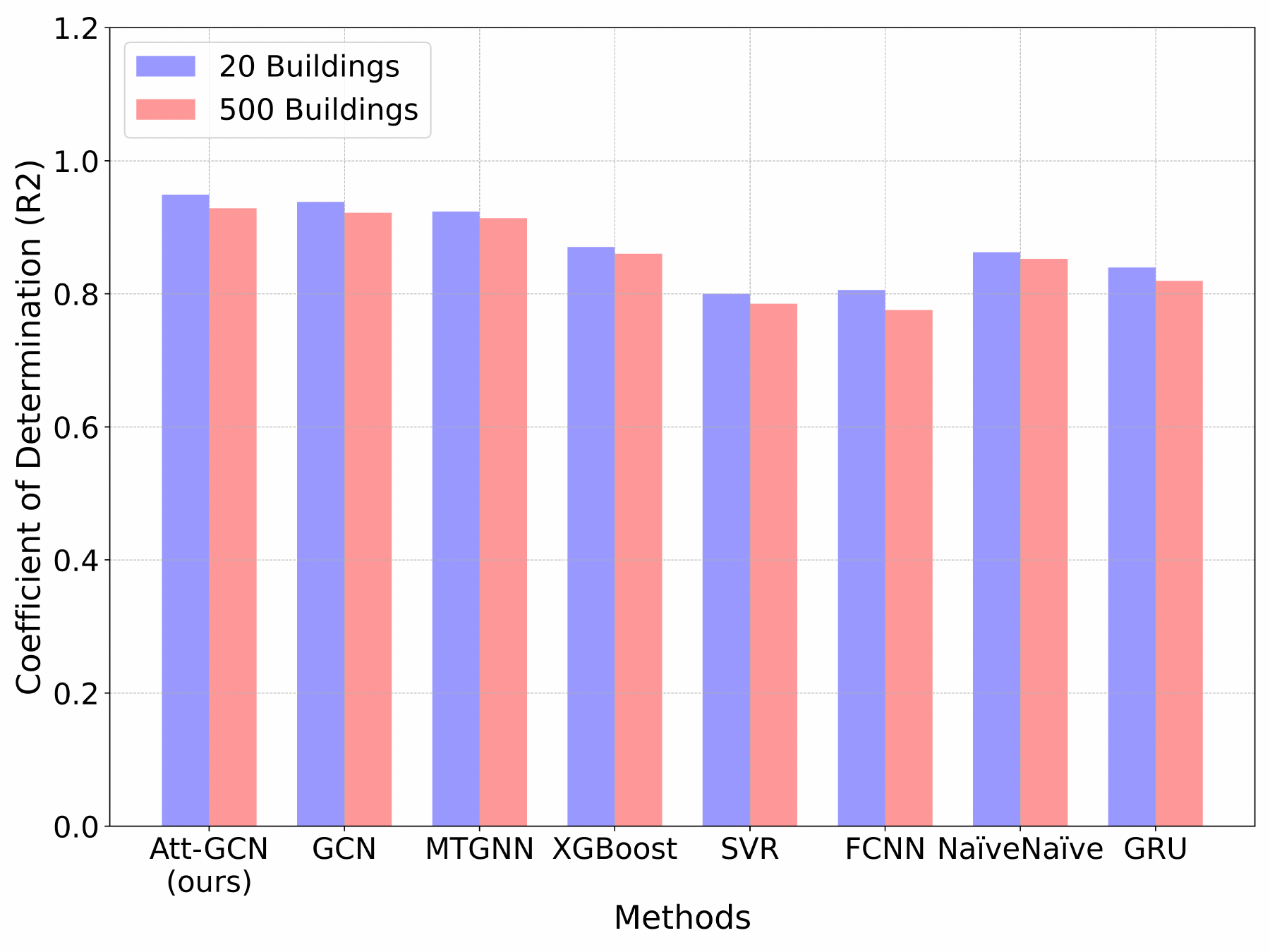} 
        \label{Fig.sub.3}} 
    \subfloat[]{ 
        \includegraphics[width=6.5cm]{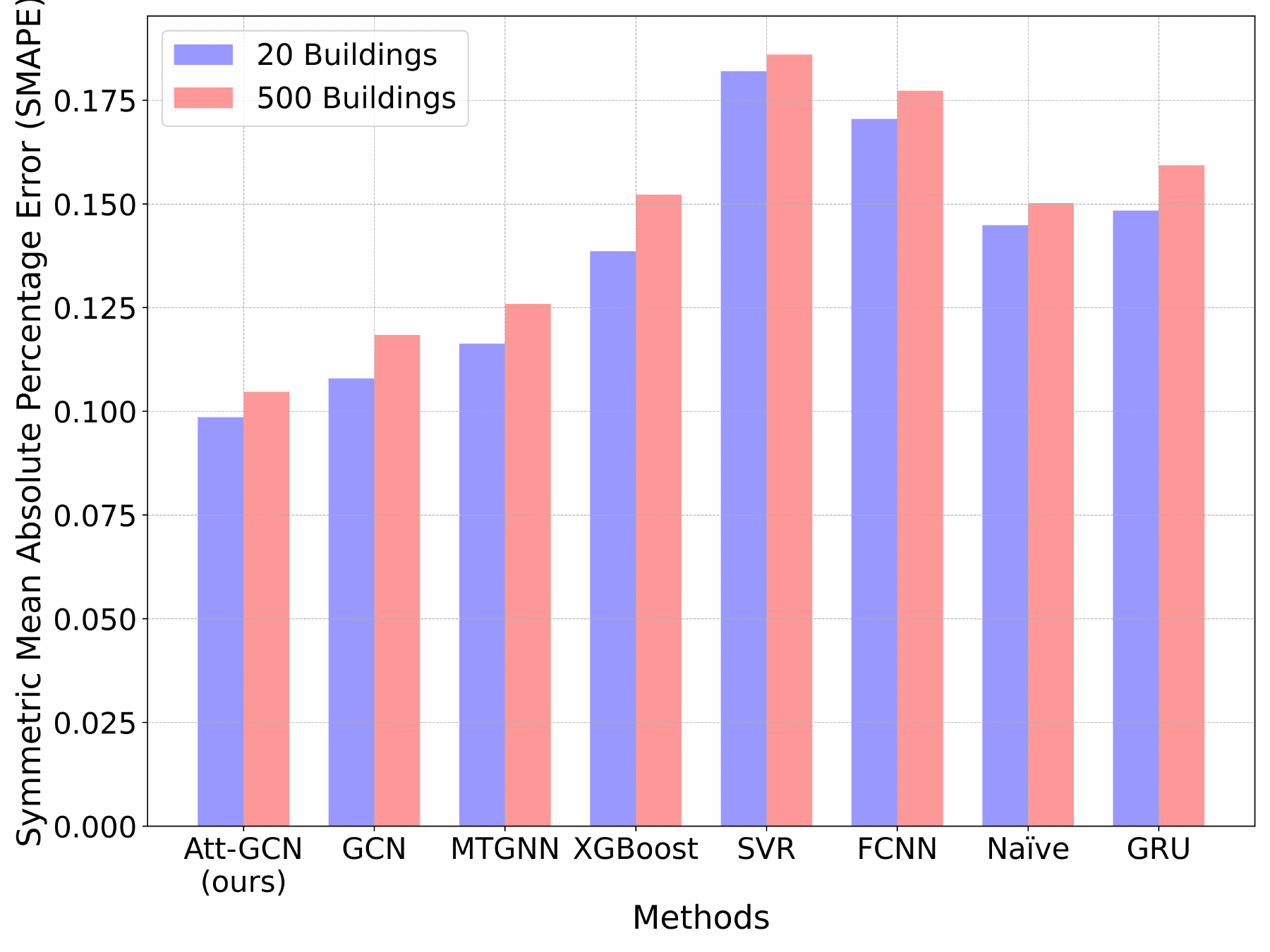} 
        \label{Fig.sub.4}}
    \caption{Prediction performance of models on the two datasets with 20 buildings and 500 buildings. (a) MSE of the load forecasting. (b) MAE of the load forecasting. (c) $R^2$ of the load forecasting. (d) SMAPE of the load forecasting.}
\end{figure*}

\subsubsection{The scalability of the mothod}
To verify the scalability of the proposed model, an evaluation was performed on the larger dataset of 500 buildings shown in the Table \ref{tab:prediction_accuracy_500}. Similarly, Att-GCN outperforms all the baselines significantly, with MSE, MAE, $R^2$ and SMAPE values of 0.0031, 0.0372, 0.9285 and 0.1047 respectively. The best method in the baseline model remained MTGNN, with MSE, MAE, $R^2$ and SMAPE of 0.0045, 0.0479, 0.9137 and 0.1259. As shown in the Fig. 3(b), Att-GCN outperforms the MSE of other models by 15.2\%(GCN), 30.6\%(MTGNN), 47.0\%(XGBoost), 59.2\%(SVR), 60.6\%(FCNN), 47.4\%(Naïve) and 55.6\%(GRU). And the MAE outperformed by 12.6\%(GCN), 22.3\%(MTGNN), 26.7\%(XGBoost), 41.4\%(SVR), 38.8\%(FCNN), 28.8\%(Naïve) and 28.9\%(GRU). The results demonstrate that the proposed method consistently and significantly improves forecasting accuracy with larger number of buildings.

In the larger dataset, compared to the results on the smaller dataset, the enhancement of Att-GCN over GCN is more pronounced, which increases from 10.3\% and 9.6\% to 15.2\% and 12.6\%. This is because the graph constructed from 500 buildings is larger, requiring deeper graph convolutional layers, and the framework incorporating attention mechanisms can effectively avoid over-smoothing, facilitating the acquisition of more accurate node representations for load forecasting. Furthermore, we analyzed the variance in predictive accuracy between two datasets. As depicted in the Fig. 5, compared to the smaller dataset, each model's MSE, MAE, $R^2$ and SMAPE on the larger dataset were slightly worse. The increase in error metrics may be attributed to the larger dataset's inclusion of a greater quantity of noise or anomalous data. Additionally, the criterion for model evaluation was the average prediction error across all buildings, which could lead to an accumulation of errors, resulting in slightly higher MSE, MAE and SMAPE, and lower $R^2$ values on the larger dataset. In summary, as the number of buildings increases, Att-GCN continues to substantially outperform other baseline models. Additionally, the attention mechanism within Att-GCN effectively prevents the over-smoothing issue commonly faced by graph convolutional networks when dealing with larger numbers of buildings. Consequently, the scalability of the model is validated.

\begin{table}[h]
    \centering
    \resizebox{\textwidth}{!}{ 
    \begin{tabular}{llllllllllll}
        \toprule
         & 0\% & 10\% & 20\% & 30\% & 40\% & 50\% & 60\% & 70\% & 80\% & 90\% & 100\%\\ \midrule
        MSE & \textbf{0.00280} & 0.00289 & 0.00291 & 0.00314 & 0.00323 & 0.00353 & 0.00378 & 0.00394 & 0.003976 & 0.003978 & 0.00406 \\
        MAE & \textbf{0.03311} & 0.03407 & 0.03469 & 0.03646 & 0.03736 & 0.03802 & 0.03944 & 0.04060 & 0.04110 & 0.04150 & 0.04317 \\
        \bottomrule
    \end{tabular}
    }
    \caption{Model robustness analysis with shuffling ratios ranging from 0\% to 100\%.}
    \label{tab:robustness}
\end{table}

\subsubsection{Robustness and generalization of the method}
Two key aspects in assessing a machine learning model's capabilities are its robustness and generalization. Robustness ensures that the model remains stable when faced with uncertainties, noise, or anomalies, while generalization gauges the model's adaptability to unseen data. Together, these traits dictate the model's usability and reliability when applied to real-world data. To evaluate the model's robustness, we introduced variability in the data by shuffling the order of the building features in the dataset at differing ratios ranging from 0\% to 100\%. The disordered building data was then input into the trained model, and the resulting load prediction accuracy are shown in the table \ref{tab:robustness}. The Fig. 6 shows that as the proportion of shuffled data increases, the MSE and MAE of the load predictions also increase, culminating in an MSE of 0.00406 and an MAE of 0.04317 at 100\% shuffling. If the metric deteriorates with the proportion of shuffled data, it indicates that the model has successfully learned spatial information. According to Fig. 6, this trend can be attributed to the disruption of the building feature matrix's order, which alters the adjacency matrix of the constructed graph, thereby influencing the final prediction results. The red line in the Fig. 6 represents the prediction accuracy of MTGNN, with MSE and MAE of 0.0042 and 0.0417, respectively. With the proportion of shuffling from 10\% to 100\%, the percentage of the increase of MSE and MAE lie in 3.58\% to 45.52\% and 2.9\% to 30.38\%. And it is evident that even under 100\% shuffling, the prediction accuracy of Att-GCN is still better than the best baseline model. Consequently, the proposed model is robust.
\begin{figure*}[t!]
    \centering
    \subfloat[]{ 
        \includegraphics[width=6.5cm]{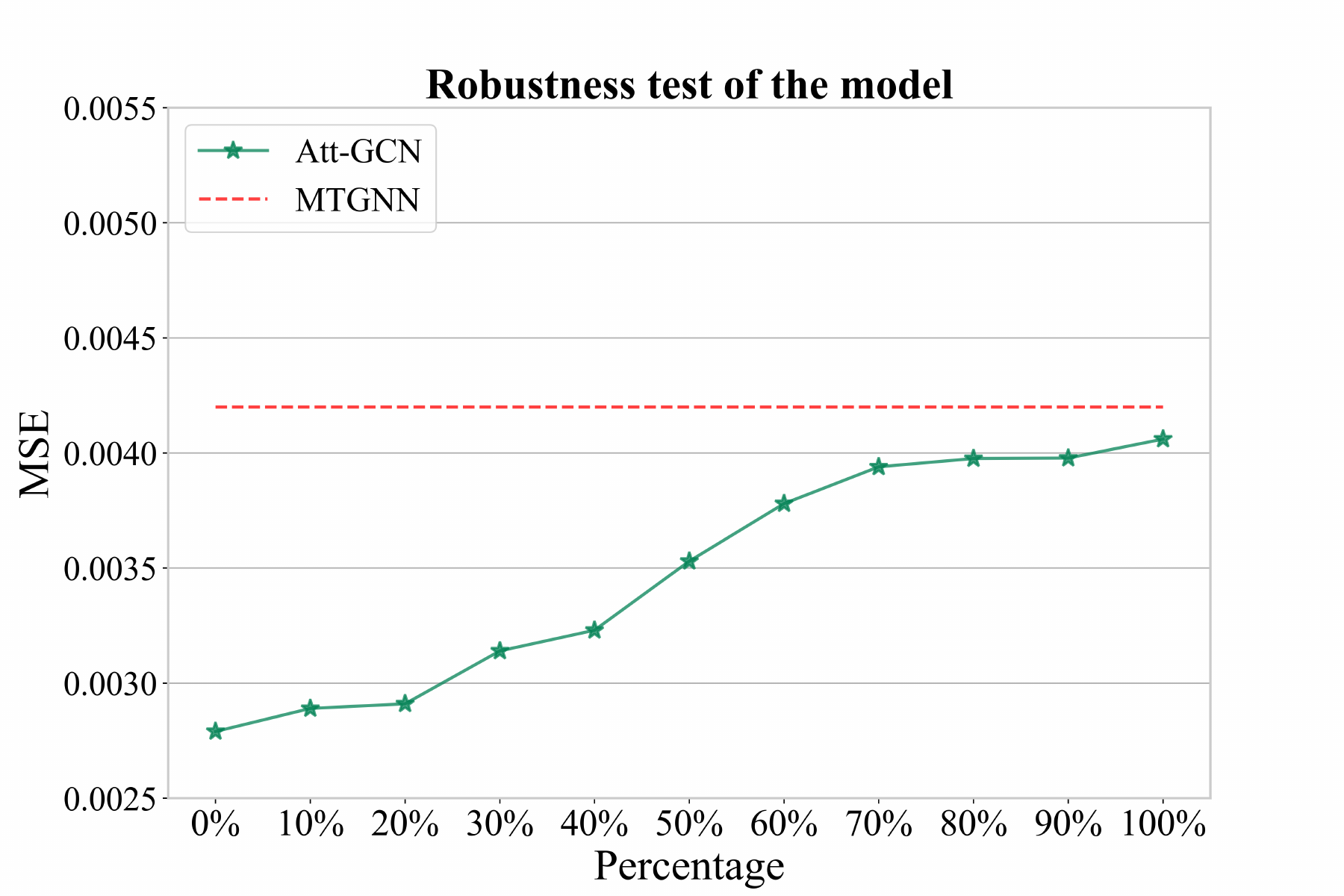} 
        \label{Fig.sub.1}} 
    \subfloat[]{ 
        \includegraphics[width=6.5cm]{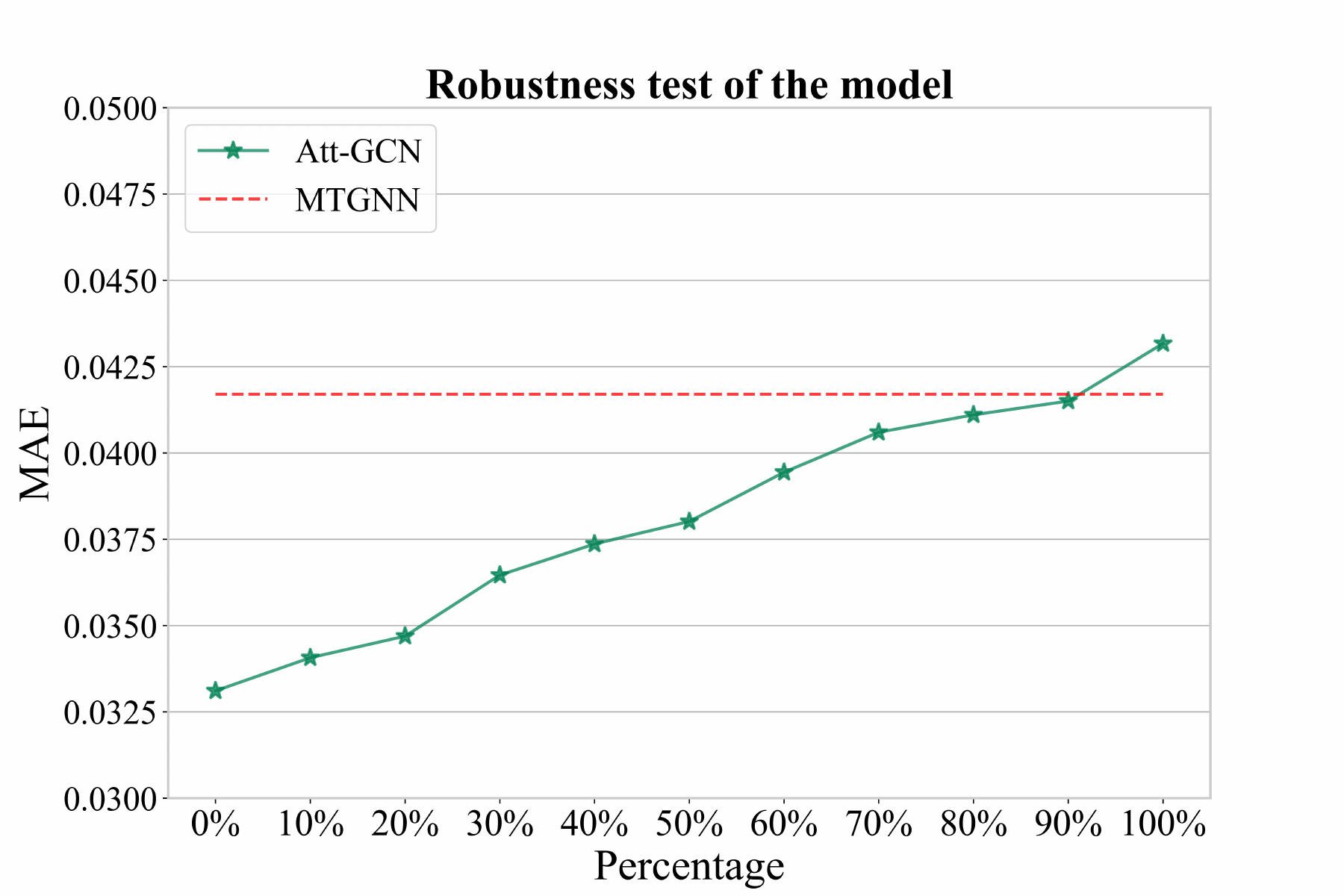} 
        \label{Fig.sub.2}} 
    \caption{Prediction load accuracy at different shuffling ratio. (a) MSE of the prediction load. (b) MAE of the prediction load.}
\end{figure*}

Additionally, the model's generalization was also evaluated on a new dataset composed of 20 buildings, with the same data format as the training dataset. The model yields a prediction load MSE of 0.0052 and an MAE of 0.0451. While we also conducted generalization tests on other baseline models using the same 20-building dataset, their performance was subpar and the MSE and MAE of the best baseline model MTGNN are 0.0083 and 0.0617. Importantly, the proposed model's prediction accuracy on this unseen dataset still outperformed the most baseline models on the test dataset, which is sufficient to demonstrate that the model's generalization is excellent. 
\begin{figure*}[t!]
    \centering
    \subfloat[]{ 
        \includegraphics[width=6.5cm]{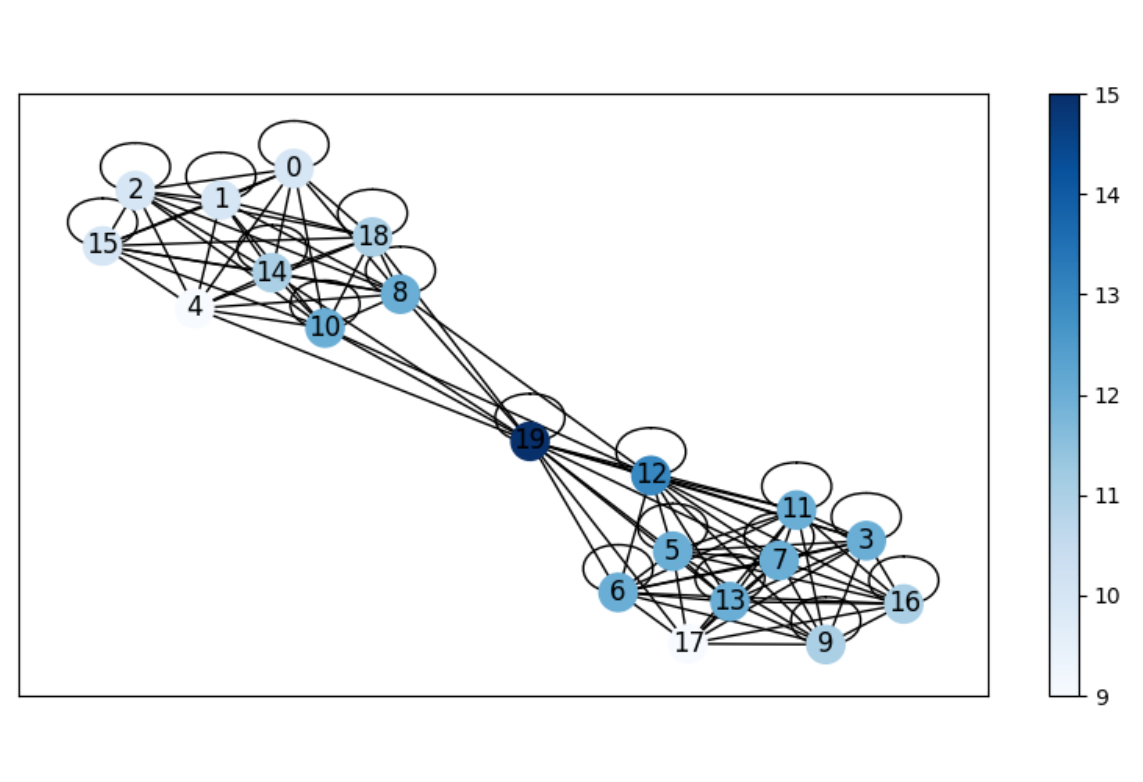} 
        \label{Fig.sub.1}} 
    \subfloat[]{ 
        \includegraphics[width=6.5cm]{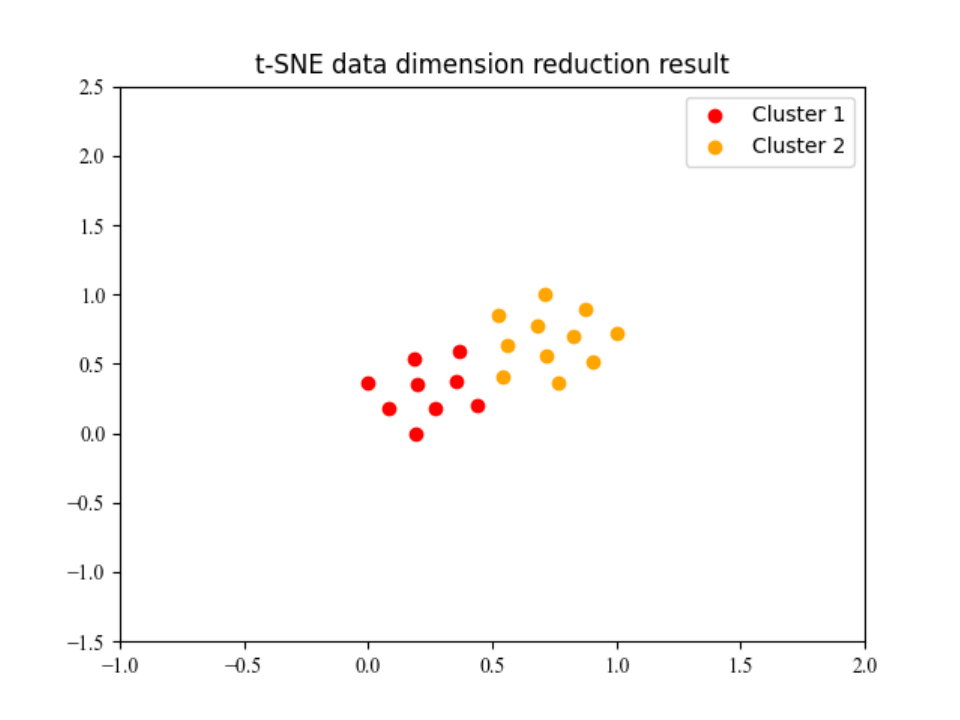} 
        \label{Fig.sub.2}} 
    \caption{Cluster results using two different methods. (a) Methods based on the connection in the graph. (b) K-means clustering.}
\end{figure*}

\subsubsection{Interpretability of the proposed method}
This section will verify the interpretability of the proposed model. Model interpretability is crucial for understanding the performance of black-box models in the field of deep learning. Specifically, in building load prediction, the interpretability of a model can greatly enhance the reliability and robustness of the system implementing the model. So the experiments on model interpretability is an essential component of the research. The graph construction method in the prediction method proposed in this research will adaptively adjust the graph structure and filter out the edges with low similarity during the model's training, so that buildings with similar energy consumption patterns form connections and cluster together in feature space. As such, the model's training process is capable of delineating the spatial dependencies among similar buildings, thus aiding load prediction.
\begin{figure*}[t!]
	\centering
	\includegraphics[width=\textwidth]{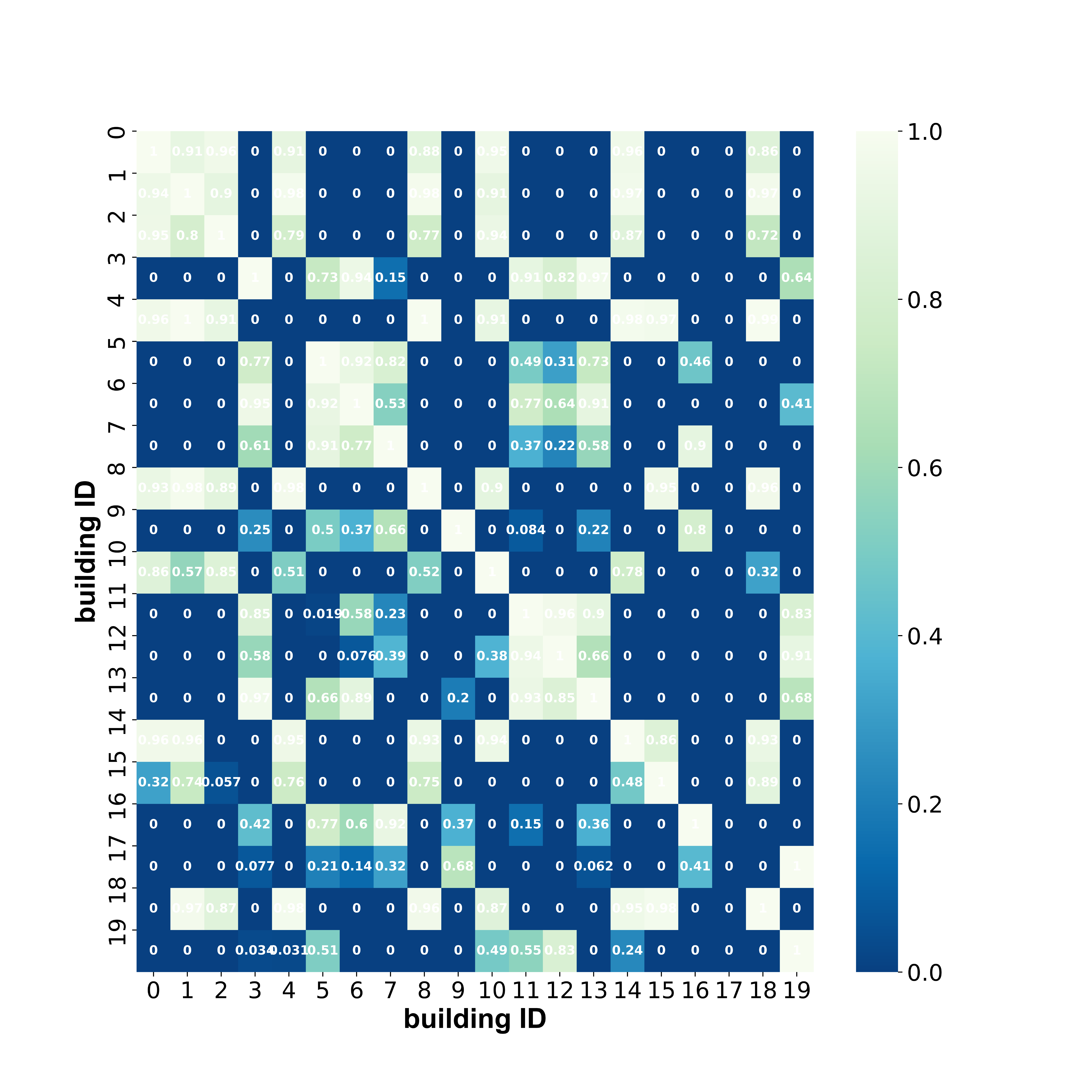} %
	\caption{The building similarity adjacency matrix.}
\end{figure*}

To validate this hypothesis, the building similarity adjacency matrix obtained from the final training output was visualized as shown in the Fig. 8, where the magnitude of the element represents the similarity between the buildings and the index on the axes represents the building number. In the adjacency matrix, zero values indicates no connection between the building nodes whereas non-zero values illustrate a connection. 

The subsequent graphical visualization of this adjacency matrix segregated the 20 buildings into two groups based on their connectivity. Meanwhile, K-means clustering was conducted on the full-year load data and outdoor environment features of the 20 buildings. To determine the optimal number of clustering groups, the silhouette coefficient \cite{rousseeuw1987silhouettes} was introduced to quantify the clustering results. The clustering results revealed that the optimal number of clusters is 2, with a silhouette coefficient of 0.658. The buildings were clustered into two groups, buildings 0, 1, 2, 4, 8, 10, 14, 15, 18 are one class and 3, 5, 6, 7, 9, 11, 12, 13, 16, 17, 19 are the other. The clustering result is visualized using the t-SNE algorithm \cite{van2008visualizing} as shown in the Fig. 7(b). Comparing the k-means clustering results with the visualization results of the adjacency matrix graph in Fig. 7(a), it can be found that the 20 buildings are all divided into two categories, and the buildings in each category correspond one-to-one. This congruency underscores the efficiency of the proposed model in identifying different types of buildings based on building features and establishing connections between similar buildings during model training. 

To further verify the rationality of the clustering, the Fig. 9 shows the load curves of the buildings within each cluster in a typical month. Although there are several difference in the magnitude of the load curves, the buildings within the same cluster exhibited noticeable homogeneity in periodicity and trends. This indicates that the buildings within the same cluster have similar energy consumption patterns, which undoubtedly validates the rationality and interpretability of of the proposed Att-GCN in learning the spatial dependencies between buildings with similar energy use patterns. 
\begin{figure*}[t!]
    \centering
    \subfloat[]{ 
        \includegraphics[width=\textwidth]{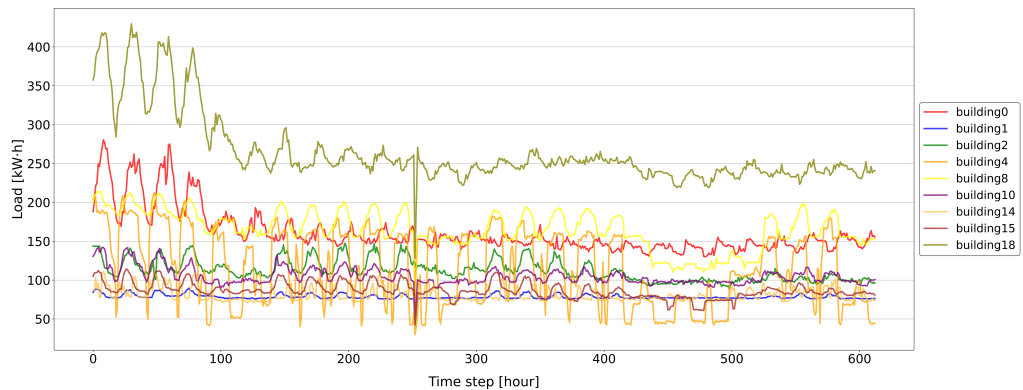} 
        \label{Fig.sub.1}} 
    \hspace{0.1em}
    \subfloat[]{ 
        \includegraphics[width=\textwidth]{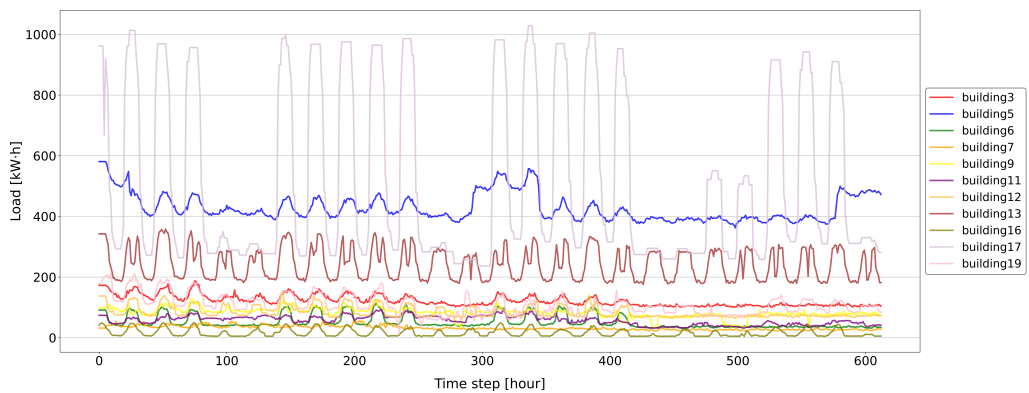} 
        \label{Fig.sub.2}} 
    \caption{Load curves of buildings in different cluster. (a) Cluster 1. (b) Cluster 2.}
\end{figure*}

%% file: 05_conclusions.tex
\section{Conclusions}
This research proposes a multi-building load forecasting method that leverages the energy interdependencies among different types of buildings, based on the spatio-temporal graph neural network. The process consists of three main steps. Firstly, the building similarity index describing the similarity between buildings is defined based on historical load and environmental factors and a building similarity adjacency matrix is constructed. This matrix serves to capture the spatial dependencies among the buildings, using a graph-based representation. The historical load data and the generated graph are then sequentially processed through Att-GCN and GRU. This step extracts the spatial dependencies between buildings and the temporal dependencies from the historical load data respectively. Finally, the building similarity adjacency matrix obtained from the model training is output to interpret the model.

In order to validate the performance of the proposed model, five typical algorithms (XGBoost, SVR, FCNN, Naïve and GRU) used for time series forecasting were selected as benchmark methods for comparison. Results revealed that the proposed model has significantly higher prediction accuracy than the other five benchmark methods. In the two datasets, MSE and MAE of the proposed model outperforms the best benchmark that did not consider the inter-building correlations (XGBoost) by 50.4\%, 31.1\% and 47.0\%, 26.7\% respectively, and outperforms the traditional GCN by 10.3\%, 50.4\%, 9.6\% and 15.2\%, 12.6\%. This suggests that the proposed model can effectively learn the spatial dependencies between buildings with similar energy usage patterns, which is beneficial for building load prediction. Additionally, the model's robustness and generalization were tested. The proposed method outperformed the baseline models on both shuffled data input and unknown datasets, demonstrating the reliability and generalization capabilities of the model. Lastly, the model's interpretability was validated. The results showed that during the training process, buildings with similar energy usage patterns were clustered together in the feature space, which is entirely consistent with the theoretical assumption of this study. 

Our research addresses the limitations of traditional building energy load prediction methods, which predominantly rely on historical operational data from individual buildings. We've expanded this approach by considering complex spatial dependencies between different types of buildings. Despite these advancements, there are several areas where our work could be further improved. Initially, this study only verified the prediction ability of the proposed model on the electrical load. It’s valuable to extend this model to other building-related applications, such as predicting cooling load, heating load, and photovoltaic power, etc. Secondly, developing our method to achieve high predictive accuracy on large-scale graphs is also a practical challenge that needs to be addressed. Besides, the study separately addresses the extraction of spatial and temporal dependencies. It’s interesting for further studies to explore an integrated approach that combines these two steps.

%% file: 06_Acknowledgment.tex
\section{Acknowledgment}
This work is financially supported by the The Major Science and Technology Projects of Ningbo under No. 2022Z236, Natural Science Foundation of Ningbo of China under No. 2023J027, China Meteorological Administration under Grant QBZ202316 as well as by the High Performance Computing Centers at Eastern Institute of Technology, Ningbo, and Ningbo Institute of Digital Twin.

%% file: 08_Appendix.tex
\appendix
\section{}

\begin{figure*}[h!]
    \centering
    \subfloat[]{ 
        \includegraphics[width=8cm]{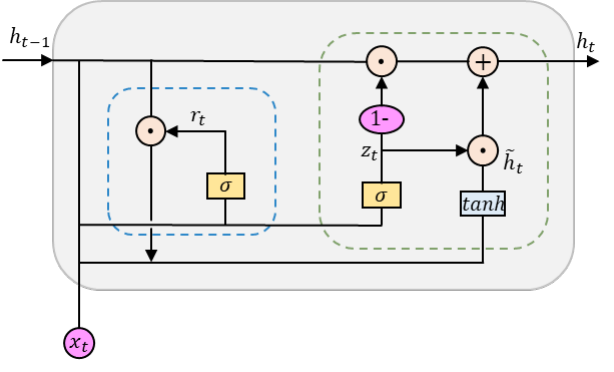} 
        \label{Fig.sub.1}} 
    \hspace{0.1em}
    \subfloat[]{ 
        \includegraphics[width=7.5cm]{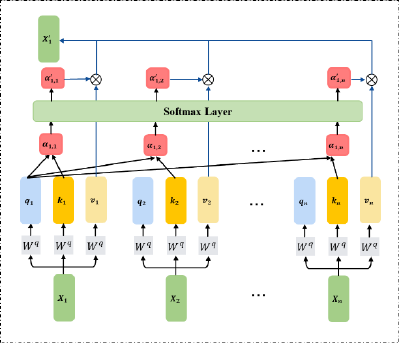} 
        \label{Fig.sub.2}} 
    \caption*{Figure A1: The structure of the related methods: (a) A GRU cell. (b) The attention mechanism.}
\end{figure*}